\documentclass[a4paper,fleqn]{cas-sc}

\usepackage[numbers]{natbib}

\usepackage{amssymb}
\usepackage{amsmath}
\usepackage{amsfonts}
\usepackage{multirow}
\usepackage{threeparttable}
\usepackage[pro]{fontawesome5}  
\usepackage{extarrows}  
\usepackage{float} 
\usepackage{subfigure} 
\usepackage{hyperref}
\usepackage{graphics}
\usepackage[title]{appendix}
\usepackage{booktabs} 

\def\tsc#1{\csdef{#1}{\textsc{\lowercase{#1}}\xspace}}
\tsc{WGM}
\tsc{QE}

\begin{document}
\let\WriteBookmarks\relax
\def\floatpagepagefraction{1}
\def\textpagefraction{.001}
\let\printorcid\relax

\shorttitle{AMF-MedIT}    

\shortauthors{Yu et al.}  

\title [mode = title]{AMF-MedIT: An Efficient Align-Modulation-Fusion Framework for Medical Image-Tabular Data}  



%
\author[a,b]{Congjing Yu}
\credit{Conceptualization, Data curation, Methodology, Validation, Writing - original draft.}
\ead{yucj@mail2.sysu.edu.cn}

\author[a,b]{Jing Ye}
\credit{Validation, Writing - review \& editing.}
\ead{yejing26@mail2.sysu.edu.cn}

\author[a,b]{Yang Liu}
\credit{Validation, Writing - review \& editing.}
\ead{liuy2529@mail2.sysu.edu.cn}

\author[c]{Xiaodong Zhang} 
\ead{dong78@smu.edu.cn}
\credit{Supervision.}

\author[a,b]{Zhiyong Zhang}\corref{cor1}
\cortext[cor1]{Corresponding author.}
\ead{zhangzhy99@mail.sysu.edu.cn}
\credit{Funding acquisition, Supervision, Writing - review \& editing.}

\affiliation[a]{organization={School of Electronics and Communication Engineering},
	addressline={Sun Yat-sen University}, 
	city={Shenzhen},
	postcode={518107}, 
	state={Guangdong},
	country={China}}
\affiliation[b]{organization={Guangdong Provincial Key Laboratory of Advanced IntelliSense Technology},
	addressline={Sun Yat-sen University}, 
	city={Shenzhen},
	postcode={518107}, 
	state={Guangdong},
	country={China}}
\affiliation[c]{organization={Department of Radiology},
	addressline={the Third Affiliated Hospital of Southern Medical University}, 
	city={Guangzhou},
	postcode={510515}, 
	state={Guangdong},
	country={China}}


\begin{abstract}
Multimodal medical analysis combining image and tabular data has gained increasing attention. However, effective fusion remains challenging due to cross-modal discrepancies in feature dimensions and modality contributions, as well as the noise from high-dimensional tabular inputs.
To address these problems, we present AMF-MedIT, an efficient Align-Modulation-Fusion framework for medical image and tabular data integration, particularly under data-scarce conditions.
Built upon a self-supervised learning strategy, we introduce the Adaptive Modulation and Fusion (AMF) module, a novel, streamlined fusion paradigm that harmonizes dimension discrepancies and dynamically balances modality contributions. It integrates prior knowledge to guide the allocation of modality contributions in the fusion and employs feature masks together with magnitude and leakage losses to adjust the dimensionality and magnitude of unimodal features.
Additionally, we develop FT-Mamba, a powerful tabular encoder leveraging a selective mechanism to handle noisy medical tabular data efficiently.
Extensive experiments, including simulations of clinical noise, demonstrate that AMF-MedIT achieves superior accuracy, robustness, and data efficiency across multimodal classification tasks. Interpretability analyses further reveal how FT-Mamba shapes multimodal pretraining and enhances the image encoder’s attention, highlighting the practical value of our framework for reliable and efficient clinical artificial intelligence applications.
\end{abstract}


%
%
%
%
%


\maketitle

\section{Introduction}
\label{sec:introduction}
Rrecently, multimodal learning that integrates medical images and tabular data has emerged as a critical research frontier in precision medicine. While computer-aided diagnosis based on images has been extensively investigated, the complementary role of tabular data in enhancing clinical decision-making is gaining attention. Such tabular data—typically derived from electronic health records (EHRs), laboratory tests, and standardized questionnaires—provide rich clinical context regarding the current symptoms and past medical history, which visual patterns alone cannot capture \cite{MZ3}. Studies have consistently shown that integrating this clinical context improves image interpretation and supports more accurate diagnoses \cite{MZ9}. 

However, real-world applications of medical images and tabular data pose two challenges. 
First, significant disparities in feature dimension and confidence levels across modalities complicate the fusion process. \textbf{(1) Dimension discrepancy:} Image encoders require high-dimensional representations to retain anatomical details, whereas tabular encoders generate more compact features. For example, a ResNet-50 encoder extracts 2,048-dimensional global features, whereas a typical feature vector dimension of TabNet is 128 \cite{TabNet}. This imbalance in feature dimensions can lead to uneven parameter allocation during fusion, where higher-dimensional features tend to dominate in parameter space. \textbf{(2) Confidence-level discrepancy:} The reliability of each modality in multimodal fusion is influenced by the downstream task and data quality. When images provide definitive pathological signatures but tabular data suffer from missing values, an optimal model is expected to prioritize image evidence. Conversely, in predicting Ischemic Heart Disease (IHD), laboratory and lifestyle factors in tabular data often serve as more reliable prognostic indicators, with images playing a supplementary role \cite{PTMS16}. This disparity in confidence necessitates fusion strategies with adaptive capabilities to appropriately balance multimodal contributions.
Second, medical tabular data inherently contain substantial noise, which hinders robust and accurate fusion \cite{TabSurvey}. These data consist of numerous features, many of which can be irrelevant to the target disease. For instance, among the selected 117 data fields for cardiac prediction in \cite{PTM2}, over 60\% have importance scores below 0.005 as measured by integrated gradients. Additionally, the prevalent problem of missing data further amplifies the noise of medical tabular data \cite{missing}.

These challenges, coupled with the persistent data scarcity in the medical domain, place greater demands on both fusion strategies and tabular representation learning, requiring a delicate balance between model capacity and data efficiency.
Although the recent adaptation of Self-Supervised Learning (SSL) to the medical field has alleviated data scarcity in encoder training, core challenges in medical image-tabular fusion and tabular feature extraction remain unresolved.
Naive concatenation fusion ignores both dimension and confidence disparities, whereas complex cross-attention mechanisms are data-hungry and computationally inefficient. 
In parallel, while architectures like Mamba \cite{Mamba} have advanced long-sequence modeling in natural language processing, medical tabular encoding still relies on either simplistic Multi-Layer Perceptrons (MLPs) or transformer-based models that require substantial training data to perform effectively.

In this work, we aim to propose an efficient fusion paradigm that resolves cross-modal disparities through a streamlined architecture, while enhancing the tabular encoder under limited data conditions. Our method is based on three insights. First, inspired by contrastive learning, constructing a unified cross-modal representation space can mitigate representation discrepancies in a straightforward manner. However, unlike matching tasks, medical inference requires preserving modality-specific distinctions \cite{PTM13}. in which only the dimensionality and magnitude of unimodal features are aligned, while allowing the contributions and essential information specific to each modality to remain distinct. Second, given the scarcity of medical multimodal data, leveraging prior knowledge from preliminary experiments enables more efficient and interpretable fusion than directly learning cross-modal interactions from limited data. Therefore, we incorporate modality confidence priors into fusion to guide adaptive contribution allocation. Third, medical tabular data can be viewed as long feature sequences containing both relevant and irrelevant elements. Mamba's selective and sequential modeling capabilities make it well-suited for processing such complex inputs \cite{Mamba}. This motivates us to explore Mamba for medical tabular representation.

Building on these insights, we present AMF-MedIT, an efficient Align–Modulation–Fusion framework for integrating medical images and tabular data. The framework first aligns the two modalities to obtain preliminary unimodal representations, then employs feature modulation to generate fusion-compatible unimodal representations, and finally fuses the modulated representations for downstream tasks.
At its core, AMF-MedIT leverages the Adaptive Modulation and Fusion (AMF) module to ensure effective fusion. The modulation aims to reduce heterogeneity among unimodal features while adaptively adjusting the contributions of each modality, enabling robust and accurate integration across diverse downstream scenarios. Specifically, we first define a modality confidence ratio $r_{\text{conf}}$ to incorporate prior information regarding modality confidence, and derive two modulation objectives. For implementation, we design a streamlined, easy-to-train, and adaptive module architecture. The modulation objectives are achieved through feature masks that truncate features, complemented by magnitude and leakage losses to guide training.

Additionally, for multimodal representation learning and alignment, we introduce FT-Mamba, an adaptation of Mamba tailored for tabular feature extraction and pretrained using contrastive learning. Leveraging the selective State-Space Mechanisms (SSMs), FT-Mamba effectively captures discriminative features from noisy medical records without relying on large-scale training data. Meanwhile, the self-supervised contrastive learning strategy enables initial multimodal alignment in an unlabeled manner, allowing FT-Mamba to provide effective supervision for the image encoder.

We conducted extensive comparative experiments on both clean and noisy datasets. Results demonstrate that AMF-MedIT achieves a favorable balance between data efficiency and multimodal performance, showing strong adaptability to noise in both image and tabular modalities. Furthermore, ablation studies and interpretability analyses highlight the respective advantages of the AMF module in fusion and FT-Mamba in tabular feature extraction.

The main contributions are summarized as follows.
\begin{itemize}
	\item We develop AMF-MedIT, an efficient framework for integrating medical images and tabular data. At its core, the AMF module is a novel modulation-based paradigm that mitigates dimension disparities and adaptively balances modality contributions by leveraging prior knowledge. 
	\item We introduce FT-Mamba, an adaptation of Mamba for medical tabular data that enhances feature extraction from noisy records, while employing self-supervised contrastive learning to enable unsupervised pretraining of image–table encoders.
	\item We conduct extensive experiments under both clean and clinically noisy conditions, showing that AMF-MedIT achieves superior accuracy, robustness, and data efficiency compared with recent methods.
	\item We provide in-depth ablation and interpretability analyses, revealing how the choice of tabular encoder influences multimodal contrastive pretraining.
\end{itemize}

\section{Related research}
\subsection{Self-Supervised Multimodal Learning}
Self-supervised learning is proposed to extract meaningful representations from unlabeled data. In multimodal learning, training models still often require expensive human annotation \cite{MZ8}. To mitigate this challenge, SSL has been applied to multimodal settings with various pretext tasks. CLIP utilizes image-text contrastive learning on massive web data, achieving remarkable zero-shot performance \cite{CLIP}. Subsequent works also combine tasks such as Masked Language Modeling (MLM) and Image-Text Matching (ITM) to enhance cross-modal understanding \cite{ALBEF}. Recently, SSL has been explored for image-tabular fusion \cite{PTM2,PTM6,PTM7,PTM13}, but existing studies have not addressed specific challenges in the medical domain. Building on contrastive learning, our approach explicitly tackles the challenges in a medical context.

\subsection{Multimodal Medical Image-Tabular Learning}
Multimodal image-tabular learning exploits tabular data to facilitate visual task learning, which is widely used in the medical field \cite{MZ3}. Several studies have employed late fusion, aggregating predictions from separate unimodal models for final decisions \cite{PTMS16, PTMS2}.
To facilitate deeper and earlier interactions between modalities, joint fusion has become more prevalent. Typically, features from different modalities are concatenated and then processed using fully connected networks \cite{PTMS17, PTMS27, CF}, transformer-based \cite{PTMS19, PTMS18}, or traditional machine learning models \cite{PTMS10} for fusion. 
To enhance modality interaction, some studies employ cross-attention modules \cite{PTMS22} to joint unimodal features. However, concatenation-based methods cannot reconcile the discrepancies between unimodal features, resulting in suboptimal fusion performance. Meanwhile, cross-attention methods rely entirely on full pairwise attention without leveraging prior knowledge, leading to high computational overhead, and making it difficult to achieve effective learning on small-scale medical datasets.

While SSL has advanced image-tabular learning, fusion strategies remain unchanged. Most studies still rely on feature concatenation \cite{PTM2, PTM7, PTM12} or cross-attention mechanisms \cite{PTM1, PTM6, PTM13, PTM15}. To address this limitation, we propose a novel fusion paradigm that flexibly regulates unimodal features within a streamlined structure, optimizing the trade-off between fusion performance and data efficiency.

\subsection{Deep Learning for Tabular Data}
Tabular data present unique challenges for deep learning (DL) models, requiring them to handle various data-related issues such as noise, imprecision, missing values, diverse attribute types, and varying value ranges \cite{TabSurvey}. Existing research on deep learning for tabular data provides diverse novel architectures. Gorishniy et al. \cite{FTT} compared the main deep architectures, including MLP, ResNet, and FT-Transformer, finding ResNet to be a simple yet effective baseline, whereas FT-Transformer serves as a universal architecture. Recently, Mamba has gained attention for leveraging selective SSMs to achieve linear-time complexity and memory efficiency \cite{Mamba}. Thielmann et al. explored its application in the tabular domain, demonstrating Mamba's feasibility for tabular data \cite{mambular}.

Despite this progress, tabular data processing in the medical domain remains relatively underdeveloped.   Some studies use raw tabular data as input \cite{PTMS10, PTMS25}, while others employ MLP-like architectures for simple processing \cite{PTMS17, PTMS18, PTMS27, PTM2}. However, such simplistic models often fall short of capturing meaningful representations from noisy and high-dimensional medical tabular data. Recent works have adopted transformer-based models to enhance tabular feature extraction \cite{PTMS19, PTM1, PTM6, PTM12, PTM13, PTM15}. Nevertheless, these models require large datasets for effective training and often encounter optimization difficulties on limited medical data.
Inspired by FT-Transformer and Mambular, we propose a streamlined adaptation of the Mamba architecture for tabular data, specifically designed to handle the noisy nature of medical tabular data.

\section{Methods}
\subsection{Overall Framework}
\begin{figure*}[h]
	\centering{\includegraphics[width=0.98\textwidth]{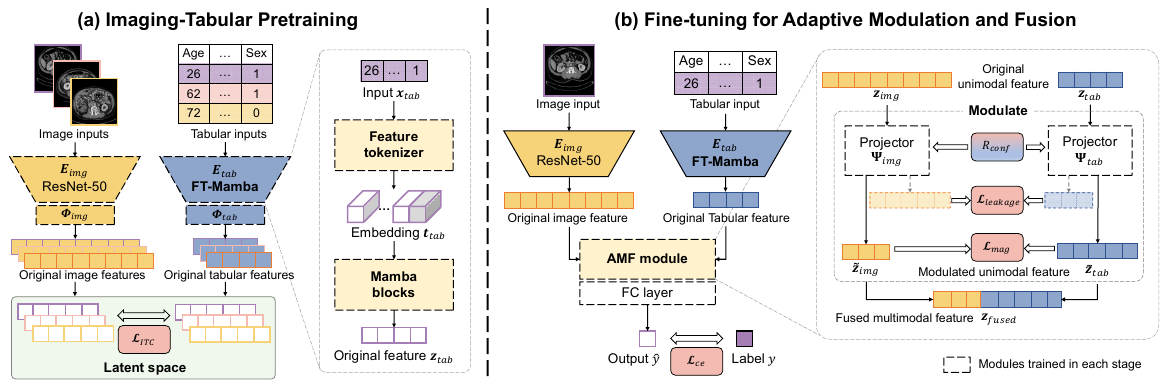}}
	\caption{The overall framework of AMF-MedIT. (a) During the contrastive pretraining stage, unimodal encoders are trained without annotations to align image and tabular representations. We introduce FT-Mamba, a powerful tabular encoder for high-dimensional, noisy medical features. (b) In the modulation and fusion stage, we propose the AMF module, a novel modulation-based paradigm that harmonizes modality feature disparities and adaptively balances their contributions via prior-guided confidence ratio $r_{\text{conf}}$, along with the magnitude and leakage loss.}
	\label{fig-overall-framework}
\end{figure*}
The overall framework of AMF-MedIT is shown in Fig. \ref{fig-overall-framework}. The training pipeline comprises two stages: (a) Imaging-tabular pretraining and (b) Fine-tuning for adaptive modulation and fusion. In each stage, the image-tabular pairs, ($\boldsymbol{x}_{img} \in \mathbb{R}^{H\times W\times D}$, $\boldsymbol{x}_{tab} \in \mathbb{R}^{N}$) are fed into the corresponding unimodal encoders, $\boldsymbol{E}_{img}$ and $\boldsymbol{E}_{tab}$. Here, $H$, $W$, and $D$ denote the height, width, and number of channels of the input image, respectively, and $N$ represents the number of tabular features. We use ResNet-50 \cite{ResNet} as the image encoder and FT-Mamba as the tabular encoder, which will be introduced in Section \ref{sec-FT-Mamba}. The encoders then generate the original unimodal embeddings, $\boldsymbol{z}_{img} \in \mathbb{R}^{D_{img}}$ and $\boldsymbol{z}_{tab} \in \mathbb{R}^{D_{tab}}$, where $D_{img}$ and $D_{tab}$ denote the dimensions of unimodal features. 

In the pretraining stage, the projection heads $\boldsymbol{\Phi}_{img}$ and $\boldsymbol{\Phi}_{tab}$ map $\boldsymbol{z}_{img}$ and $\boldsymbol{z}_{tab}$ into a shared latent space as $\boldsymbol{\hat{z}}_{img} \in \mathbb{R}^{D_{\Phi}}$ and $\boldsymbol{\hat{z}}_{tab} \in \mathbb{R}^{D_{\Phi}}$ separately. $D_{\Phi}$ denotes the dimensions of projections.  On the supervise of the Image-Tabular Contrastive (ITC) loss, the projections are pulled and pushed to maximize the similarity between matching image-tabular pairs while minimizing the similarity of non-matching pairs.

In the fine-tuning stage, we propose a simple AMF module for multimodal fusion. Guided by the proposed magnitude and leakage losses, this module transforms the original unimodal features, $\boldsymbol{z}_{img}$ and $\boldsymbol{z}_{tab}$, into modulated features, $\boldsymbol{\tilde{z}}_{img}$ and $\boldsymbol{\tilde{z}}_{tab}$, which are then combined to form a cohesive fused feature, $\boldsymbol{z}_{fused}$. Finally, $\boldsymbol{z}_{fused}$ is fed to a Fully-Connected (FC) layer to generate the classification probability.

\subsection{FT-Mamba}
\label{sec-FT-Mamba}
Inspired by FT-Transformer \cite{FTT}, we adapt the Mamba architecture \cite{Mamba} for tabular feature extraction and name it as FT-Mamba (Feature Tokenizer + Mamba). As shown in Fig. \ref{fig-overall-framework} (a), we first separately tokenize the categorical and numerical tabular features to embeddings and add a \texttt{[CLS]} token at the sequence end. Then a stack of mamba blocks is used to extract the high-dimensional features from the embeddings, and the final representation of the \texttt{[CLS]} token is used as the original tabular embedding, $\boldsymbol{z}_{tab}$.
\begin{figure}[h]
	\centering{\includegraphics[width=0.6\textwidth]{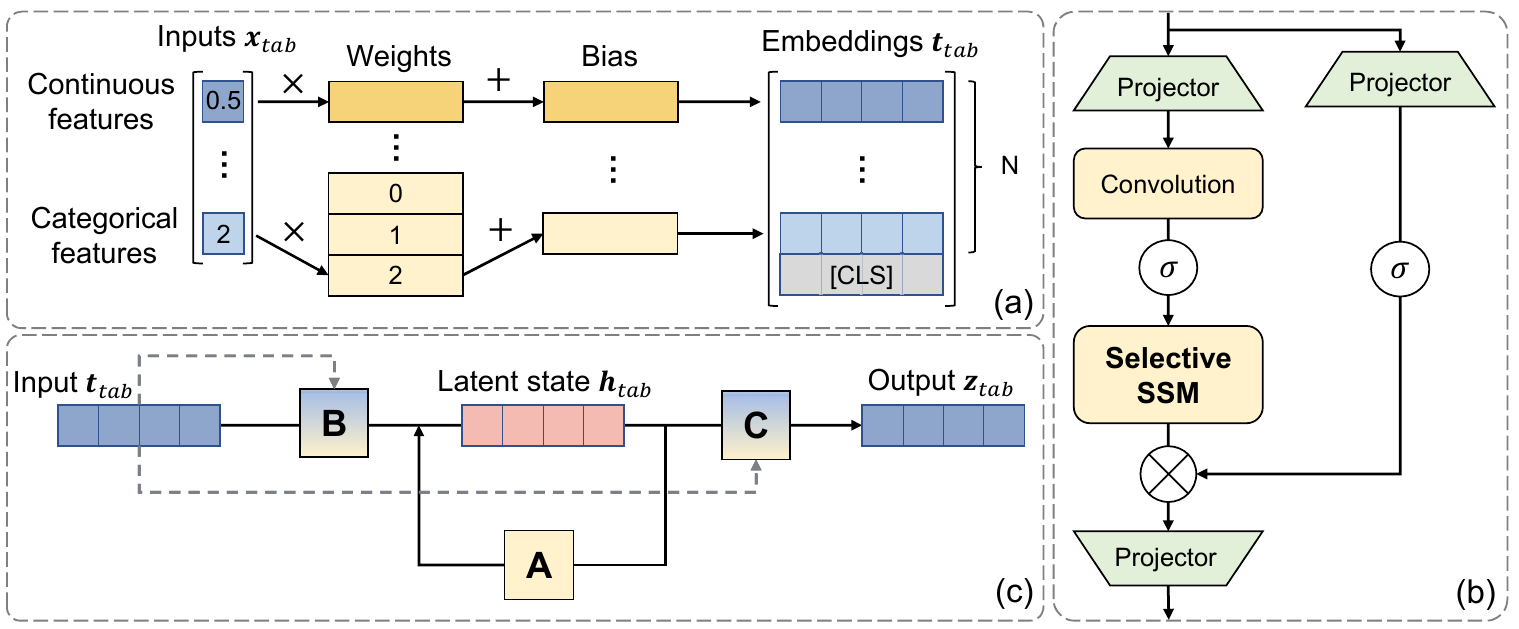}}
	\caption{The key modules in FT-Mamba. (a) Feature tokenizer: categorical and numerical tabular features are separately tokenized into embeddings; (b) a Mamba block, and its core module (c) Selective SSM: it enables input-dependent parameterization of SSMs and dynamically adjusts them to focus on relevant input features.}
	\label{fig-FT-Mamba}
\end{figure}

\subsubsection{Feature Tokenizer} 
Our feature tokenizer module follows the design of \cite{FTT}. It transfers the tabular input $\boldsymbol{x}_{tab}$ to embeddings $\boldsymbol{t}_{tab} \in \mathbb{R}^{N \times D_{tab}}$ as shown in Fig. \ref{fig-FT-Mamba} (a). For continuous features, the feature value is mapped by a linear function with $D_{tab}$-dimensional weights and bias. For categorical features, each category is mapped to a vector through a lookup table and is added with the corresponding bias.

\subsubsection{Mamba Blocks} 
As shown in Fig. \ref{fig-FT-Mamba} (b) and (c), Mamba maps input token ${\boldsymbol{t}_{tab}}$ to the representation ${\boldsymbol{z}_{tab}}$ through a hidden state ${\boldsymbol{h}_{tab}}$, which stores and updates long-term information without explicit self-attention. Moreover, the module employs adaptive parameters $\mathbf{B}, \mathbf{C}$ as functions of the input, allowing it to selectively focus on relevant information while filtering out irrelevant features. Unlike transformers, which compute all pairwise interactions, Mamba reduces unnecessary computations, facilitating efficient optimization in data-limited scenarios.

\subsection{Adaptive Modulation and Fusion Module}
\subsubsection{Problem Formulation}
\label{s3.3.1}
Modality feature heterogeneity and imbalanced modality contributions can undermine the effectiveness of multimodal fusion. To address this, feature modulation aims to make the original unimodal features more compatible for fusion by mitigating heterogeneity in feature magnitude and dimensionality across modalities, while leveraging prior knowledge regarding modality confidence to adjust their contributions.

We first employ the $\ell_1$-norm to quantify the contribution of a unimodal feature in naive multimodal concatenation. The $\ell_1$-norm of a feature element directly reflects its magnitude, and the sum over all elements provides a simple and interpretable measure of the overall contribution $\mathcal{C}$ of a unimodal feature, as in \eqref{overall-contribution}. Intuitively, modalities with larger overall contributions dominate in concatenated representations.
\begin{subequations}
	\begin{align}
	\mathcal{C}(\boldsymbol{z}_{img}) &\xlongequal{\text{def}} {\left\|\boldsymbol{z}_{img}\right\|}_1 = \sum_{i=0}^{D_{img}}\left|{z_{img}}_{i}\right|
	\label{overall-contribution}\\
	\bar{\mathcal{C}}(\boldsymbol{z}_{img}) &\xlongequal{\text{def}} \frac{{\left\|\boldsymbol{z}_{img}\right\|}_1}{D_{img}} = \frac{\sum_{i=0}^{D_{img}}\left|{z_{img}}_{i}\right|}{D_{img}}
	\label{average-contribution}
	\end{align}
\label{contribution}
\end{subequations}
To eliminate the influence of dimensionality, we define the average contribution $\bar{\mathcal{C}}$ as in \eqref{average-contribution}, which reflects the average magnitude per feature element. 
Equations \eqref{overall-contribution} and \eqref{average-contribution} use $\boldsymbol{z}_{img}$ as an example, and the same concept applies to the tabular modality. 

To further incorporate prior knowledge regarding modality confidence into fusion, we introduce the modality confidence ratio $ r_{\text{conf}}$. This ratio represents the confidence level of image features relative to tabular features or a given downstream task, and the confidence levels can be derived from prior knowledge, such as preliminary experiments or expert experience. In our experiments, $ r_{\text{conf}}$ is calculated based on unimodal performance metrics obtained from pretrained encoders on downstream tasks, denoted as $\textbf{Metrics}$:
\begin{equation}
	r_{\text{conf}} = \frac{\textbf{Metrics}(\boldsymbol{z}_{img})}{\textbf{Metrics}(\boldsymbol{z}_{tab})}. 
	\label{Rconf}
\end{equation} 

Based on the above definitions, we propose two objectives for modulation:
\begin{itemize}
	\item[1.] \textbf{Magnitude alignment.} To avoid the large-magnitude features overwhelming the smaller ones, the average feature magnitude across modulated unimodal features should be comparable:
\begin{equation}
	\left|\bar{\mathcal{C}}(\boldsymbol{\tilde{z}}_{img}) - \bar{\mathcal{C}}(\boldsymbol{\tilde{z}}_{tab}) \right| \rightarrow 0.
	\label{require:magnitude}
\end{equation} 
	\item[2.] \textbf{Confidence and dimensionality adjustment.} To ensure that more reliable modalities contribute proportionally more to downstream tasks, the overall contribution of each modality in the fused representation should be proportional to its confidence level. Given the constraint of comparable feature magnitudes as required in \eqref{require:magnitude}, the overall contribution of a modality is mainly determined by its feature dimensionality according to \eqref{average-contribution}. Hence, the relative contributions can be modulated by scaling the feature lengths, which can be readily implemented within neural networks. This objective can be expressed as:
	\begin{equation}
		\frac{\mathcal{C}(\boldsymbol{z}_{img})}{\mathcal{C}(\boldsymbol{z}_{tab})} = r_{\text{conf}} \Rightarrow \frac{L_{img}}{L_{tab}} = r_{\text{conf}}, 
		\label{require:length}
	\end{equation} 
	where $L_{img}$ and $L_{tab}$ denote the dimension of $\boldsymbol{\tilde{z}}_{img}$ and $\boldsymbol{\tilde{z}}_{tab}$, respectively. 
\end{itemize}
	
To achieve the two objectives in \eqref{require:magnitude} and \eqref{require:length}, we propose the AMF module, which modulates the original unimodal features using a straightforward architecture and simple training tasks, enabling effective and interpretable multimodal fusion. 
	
\subsubsection{Implementation}
\begin{figure*}[pos=h]
	\centering{\includegraphics[width=0.6\textwidth]{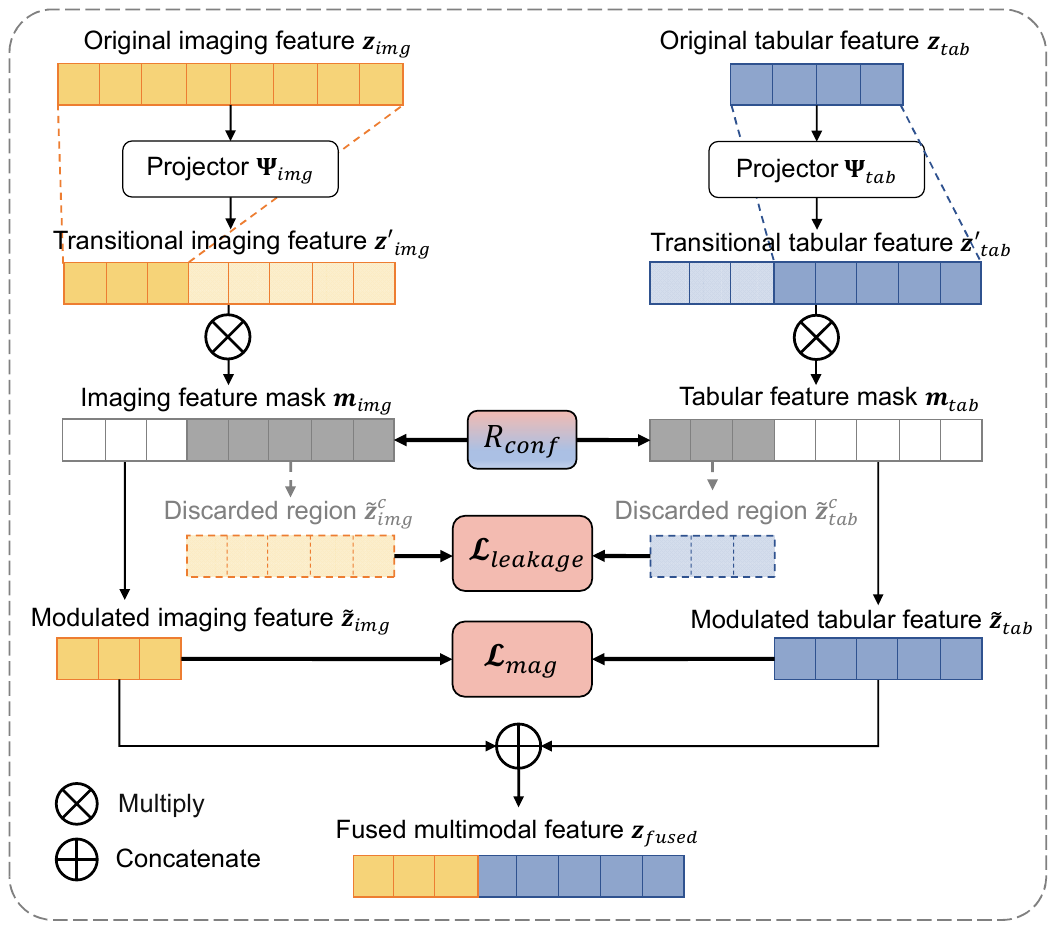}}
	\caption{The structure of the AMF module. For each modality, the original features are mapped to the transitional features separately. Then feature masks implicitly partition them into retained region (solid lines) and discarded region (dashed lines). During training, the leakage loss compresses effective information into the retained region, and magnitude loss aligns their magnitude across modalities.}
	\label{fig-AMFM}
\end{figure*}

The structure of the AMF module is shown in the Fig. \ref{fig-AMFM}. First, two projector heads, $\boldsymbol{\Psi}_{img}$ and $\boldsymbol{\Psi}_{tab}$, map the original features $\boldsymbol{z}_{img}$ and $\boldsymbol{z}_{tab}$ to the transitional features $\boldsymbol{z}'_{img}, \boldsymbol{z}'_{tab} \in \mathbb{R}^{D_{out}}$, where $D_{out}$ corresponds to the dimension of the fused feature. Then, the adjustment of feature dimensions according to $r_{\text{conf}}$ is not realized by changing the structure of the projection heads. Instead, it is realized by learning appropriate parameter values during training, guided by the subsequent feature mask and leakage loss. This design enhances the stability and generalizability of the module: when $r_{\text{conf}}$ varies due to changes in downstream data quality, there is no need to modify the projection head structure, as the AMF module can adaptively modulate the features.
Finally, the modulated features, $\boldsymbol{\tilde{z}}_{img} \in \mathbb{R}^{L_{img}}$ and $\boldsymbol{\tilde{z}}_{tab} \in \mathbb{R}^{L_{tab}}$, are concatenated to obtain the fused feature $\boldsymbol{z}_{fused}$ with a dimension of $D_{out} = L_{img} + L_{tab}$, which is subsequently fed into the decoder for downstream tasks. 

\paragraph{\textbf{Feature mask.}}
In practical implementation, we fix $D_{out}$ as a hyperparameter to ensure a consistent interface to downstream decoders. This design decouples the adjustment of modality contributions from the architecture of the fusion and decoding modules, thereby improving the generality of the AMF module and avoiding repeated reconfiguration of decoder input layers. 
Given a fixed $D_{out}$ and combined with \eqref{require:length}, the allocation of dimensions between modalities is adaptively determined by the modality confidence ratio $r_{\text{conf}}$ as follows:
\begin{equation}
	D_{out} \;\text{is fixed}, \quad 
	\begin{cases}
		L_{img}(r_{\text{conf}}) &= D_{out} - \left\lfloor \frac{D_{out}}{1+r_{\text{conf}}} \right\rfloor, \\
		L_{tab}(r_{\text{conf}}) &= \left\lfloor \frac{D_{out}}{1+r_{\text{conf}}} \right\rfloor.
	\end{cases}
	\label{length}
\end{equation}

To explicitly adjust the dimensionalities of unimodal features, we define a feature mask for each modality to truncate the transitional features. For the imaging modality, the mask is a binary vector of size $D_{out}$, where the first $L_{img}$ entries are set to $1$ and the remaining $L_{tab}$ entries are set to $0$. Conversely, the tabular feature mask adopts the opposite configuration:
\begin{equation}
	\label{mask}
	\begin{cases}
		\boldsymbol{m}_{img} &= [\underbrace{1, 1, \dots, 1}_{L_{img}}, \underbrace{0, 0, \dots, 0}_{L_{tab}}] \\ \boldsymbol{m}_{tab} &= [\underbrace{0, 0, \dots, 0}_{L_{img}}, \underbrace{1, 1, \dots, 1}_{L_{tab}}].
	\end{cases}
\end{equation}
By applying the feature masks to the transitional features, the features are divided into a retained region, $\tilde{\boldsymbol{z}}_*$, and a discarded region, $\tilde{\boldsymbol{z}}_{*}^{c}$, as formulated in \eqref{mask}. Consequently, the retained region of each modality satisfies the dimensionality objective specified in \eqref{require:length}. 
\begin{equation} 
	\label{mask} 
	\begin{aligned}
		\tilde{\boldsymbol{z}}_{img} &= \left[\boldsymbol{m}_{img} \odot \boldsymbol{z}'_{img}\right]_{0:L_{img}}, &\quad \tilde{\boldsymbol{z}}^c_{img} &= \left[\boldsymbol{m}_{tab} \odot \boldsymbol{z}'_{img}\right]_{L_{img}:D_{out}};\\
		\tilde{\boldsymbol{z}}_{tab} &= \left[\boldsymbol{m}_{tab} \odot \boldsymbol{z}'_{tab}\right]_{L_{img}:D_{out}}, &\quad
		\tilde{\boldsymbol{z}}^c_{img} &= \left[\boldsymbol{m}_{img} \odot \boldsymbol{z}'_{tab}\right]_{0:L_{img}}
	\end{aligned}
\end{equation}
Here, $\odot$ denotes the element-wise product between two vectors. During training, the projections are optimized under the guidance of the leakage and magnitude losses, which encourage modality-specific information to be concentrated in the retained region while maintaining comparable feature magnitudes. The retained regions are then used as the modulated features, which are concatenated to form the final fused representation as follows:
\begin{equation} 
	\label{concat} 
		\boldsymbol{z}_{fused} = {\boldsymbol{m}_{img}} \odot \boldsymbol{z}'_{img} + {\boldsymbol{m}_{tab}} \odot \boldsymbol{z}'_{tab}.
\end{equation}

\paragraph{\textbf{Leakage loss.}}
A potential issue with masking is that meaningful information may “leak” into the discarded region, thereby reducing the effectiveness of the retained features. To address this, we introduce a leakage loss to measure the residual information magnitude within the masked-out regions:
\begin{equation} 
	\label{leakage} 
	\mathcal{L}_{\mathrm{leakage}}
	= \frac{1}{2\mathcal{N}}\sum_{j\in\mathcal{N}}
	\left(
	\bar{\mathcal{C}}\big(\tilde{\boldsymbol{z}}_{img, j}^{c}\big)
	+ \bar{\mathcal{C}}\big(\tilde{\boldsymbol{z}}_{tab, j}^{c}\big)
	\right)
	= \frac{1}{2\mathcal{N}}\sum_{j\in\mathcal{N}}
	\left(
	\frac{\left\|\left[\boldsymbol{z}'_{img, j}\right]_{L_{img}:D_{out}}\right\|_1}{L_{tab}}
	+
	\frac{\left\|\left[\boldsymbol{z}'_{img, j}\right]_{0:L_{img}}\right\|_1}{L_{img}}
	\right).
\end{equation}
where $\mathcal{N}$ is the batch size. By minimizing this loss jointly with the cross-entropy (CE) loss, the projection heads $\boldsymbol{\Psi}_{img}$ and $\boldsymbol{\Psi}_{tab}$ are encouraged to compress discriminative information into the retained region, ensuring that truncation by feature masks does not compromise the expressiveness of the fused features.

\paragraph{\textbf{Magnitude loss.}}
To satisfy the objectives in \eqref{require:magnitude}, we define a magnitude loss that measures the discrepancy in information magnitude between the retained feature regions of the two modalities:
\begin{equation}
	\label{magnitude-loss}
	\mathcal{L}_{\text{mag}} = \frac{1}{\mathcal{N}} \sum_{j\in \mathcal{N}} \left(\bar{\mathcal{C}}(\tilde{\boldsymbol{z}}_{img, j}) - \bar{\mathcal{C}}(\tilde{\boldsymbol{z}}_{tab, j}) \right) 
	= \frac{1}{\mathcal{N}} \sum_{j\in \mathcal{N}} \left| \frac{{\left\| \left[{\boldsymbol{z}'_{img, j}}\right]_{0:L_{img}} \right\|}_1}{L_{img}} - \frac{{\left\| \left[{\boldsymbol{z}'_{tab,j}}\right]_{L_{img}:D_{out}} \right\|}_1}{L_{tab}} \right|.
\end{equation}
By minimizing this loss, the module is guided to reduce the magnitude gap across modalities, while ensuring the feasibility of adjusting the overall modality contribution through dimensional modulation.

\subsection{Optimization}
In the pretraining stage, image–tabular pairs are used for contrastive learning and the ITC loss is applied \cite{PTM2, CLIP}. Intuitively, the goal of ITC is to align paired image and tabular representations while pushing apart unpaired ones. The image feature of a given sample is encouraged to be close to its corresponding tabular feature, and simultaneously distant from tabular features of other samples in the batch. 
Formally, the image-to-tabular contrastive loss is defined in \eqref{CLIP-imaging-loss}, where $\mathcal{N}$ is the batch size and $\tau$ is the temperature hyperparameter:
\begin{equation}
	\mathcal{L}_{\text{i2t}} = -\frac{1}{\mathcal{N}}\sum_{j\in \mathcal{N}} log\frac{\text{exp}(\text{cos}({\boldsymbol{\hat{z}}_{img}}^j, {\boldsymbol{\hat{z}}_{tab}}^j)/\tau)} {\sum\limits_{k\in \mathcal{N},k\neq j}\text{exp}(\text{cos}({\boldsymbol{\hat{z}}_{img}}^j, {\boldsymbol{\hat{z}}_{tab}}^k)/\tau)} \\
	\label{CLIP-imaging-loss}
\end{equation}
The tabular-to-image contrastive loss adopts a symmetric formulation, and the overall ITC loss is computed as their average as: $\mathcal{L}_{\text{ITC}} = \left(\mathcal{L}_{\text{i2t}} + \mathcal{L}_{\text{t2i}} \right)/2.$

In the fine-tuning stage, the loss function consists of three components. Alongside the leakage loss in \eqref{leakage} and magnitude loss in \eqref{magnitude-loss} for multimodal fusion, we employ CE loss for classification. For the OL3I dataset, which suffers from severe class imbalance, we apply a weighted CE loss as defined in \eqref{wce-loss}, where $p$ is the model output, $y$ is the classification label, and $\alpha$ is a hyperparameter for mitigate class imbalance. 
\begin{equation}
	\label{wce-loss}
	\mathcal{L}_{\text{weighted ce}} = -\frac{1}{\mathcal{N}} \sum_{j\in \mathcal{N}} \left( \alpha y^j log(p^j)-(1-\alpha)(1-y^j)log(1-p^j) \right)
\end{equation}
For the DVM dataset, standard CE loss is used as formulated in \eqref{ce-loss}, where $\mathcal{C}$ denotes the number of classes.
\begin{equation}
	\label{ce-loss}
	\mathcal{L}_{\text{ce}} = -\frac{1}{\mathcal{N}} \sum_{j\in \mathcal{N}} \sum_{c\in \mathcal{C}}   y_c^j log(p_c^j)
\end{equation}
The overall loss function is a weighted sum of these three losses, with $\lambda_1, \lambda_2$ as hyperparameters to balance leakage and magnitude loss as follows:
\begin{subequations}
	\label{fine-tuning-loss}
	\begin{align}
		\mathcal{L}_{\text{OL3I}} &= \mathcal{L}_{\text{weighted ce}} + \lambda_1\mathcal{L}_{\text{leakage}} + \lambda_2\mathcal{L}_{\text{magnitude}} \\
		\mathcal{L}_{\text{DVM}} &= \mathcal{L}_{\text{ce}} + \lambda_1\mathcal{L}_{\text{leakage}} + \lambda_2\mathcal{L}_{\text{magnitude}}.
	\end{align}
\end{subequations}

\section{Experiments}
\subsection{Datasets}
We conducted experiments on two multimodal datasets: OL3I (Opportunistic L3 computed tomography slices for Ischemic Heart Disease risk assessment) \cite{OL3I} and DVM (Data Visual Marketing) \cite{DVM}. OL3I is a medical dataset that combines grayscale computed tomography (CT) images with high-dimensional clinical tabular features, while DVM is a large-scale natural dataset for automotive applications, offering a complementary benchmark for general-purpose scenarios.

The OL3I dataset contains 8,139 axial CT slices at the third lumbar vertebra (L3) level, each paired with structured medical record data. Every image is a 512$\times$512 grayscale slice, and the tabular modality includes 423 features covering patient demographics, laboratory results, and aggregated clinical codes. In our study, we adopted the 1-year IHD prediction task as the downstream objective. Due to the severe class imbalance, we used the Area Under the receiver operating Characteristic Curve (AUC) as the evaluation metric. All experiments followed the dataset’s predefined training, validation, and test splits.

The DVM dataset comprises 35,719 image-tabular pairs, each consisting of a 300$\times$300 front-view RGB image of a car and corresponding tabular data on its sales and technical data. To avoid label leakage in the car model classification task, we excluded features such as brand, model year, and vehicle size, resulting in 13 tabular input features. Following the protocol in \cite{PTM2}, we considered 101 target classes, each with at least 100 samples. Top-1 accuracy (ACC) was used as the evaluation metric, and the data was randomly split into training, validation, and test sets with a ratio of 72\%, 18\%, and 10\%, respectively. Additional preprocessing details are provided in Appendix~\ref{AA1}.

\subsection{Experimental Setup}
In our experiments, the image encoder is a ResNet-50, which generates image features of dimension 2048. The tabular encoder is implemented as FT-Mamba, consisting of two Mamba blocks, and produces tabular features of dimension 64. All projection heads in the network are MLPs with a single hidden layer. During pretraining, the projection heads map features to a 128-dimensional space \cite{PTM2}. In the AMF module, the projection heads project features to a 2048-dimensional space. A fully connected layer is used for the final classification.

We first pretrain the image and tabular encoders on the OL3I and DVM datasets, respectively, following the contrastive learning strategy described in \cite{PTM2}. Prior to fine-tuning, each pretrained encoder is frozen and paired with an FC classification layer to form a unimodal model. The unimodal performance on the downstream task is then used to compute the modality confidence ratio $r_{\text{conf}}$ according to \eqref{Rconf}. For downstream tasks, fine-tuning is performed under two strategies: in the frozen strategy, the parameters of pretrained encoders remain fixed, whereas in the trainable strategy, all parameters are updated during training. All results are reported as the mean and standard deviation across five different random seeds, and more implementation details are provided in Appendix~\ref{AA}.

\subsection{Comparison with Recent Methods on Downstream Classification}
To evaluate the effectiveness of our proposed AMF-MedIT framework, we compare it with both supervised baselines and recent SSL approaches on the two downstream classification tasks. The supervised baselines include a ResNet-50 for the image modality \cite{ResNet} , an FT-Transformer for the tabular modality \cite{FTT}, and a Concat Fuse method for medical image and tabular fusion \cite{CF}. For SSL-based multimodal methods, we include MMCL \cite{PTM2}, TIP \cite{PTM6}, and AMF-MedIT. MMCL and AMF-MedIT follow the same pretraining strategy, while TIP adopts the pretraining procedure described in its original paper.
\begin{table*}[pos=h]
	\centering
	\caption{Classification performance of AMF-MedIT and recent methods for image and tabular data.}
	\label{Multi-modal}
	\begin{threeparttable}
			\begin{tabular}{l | cc | cc}
				\toprule
				\multirow{2}{*}{\textbf{Model}} & 
				\multicolumn{2}{c|}{\textbf{OL3I AUC (\%)}} & 
				\multicolumn{2}{c}{\textbf{DVM ACC (\%)}} \\
				\cmidrule(lr){2-3} \cmidrule(lr){4-5}
				& \faSnowflake\tnote{1} & \faFire*\tnote{2} & \faSnowflake & \faFire* \\
				\midrule
				\multicolumn{5}{c}{\textbf{Supervised Unimodal and Multimodal Methods}} \\
				\midrule
				ResNet-50 \cite{ResNet}        & \multicolumn{2}{c|}{58.29 $\pm$ 4.16} & \multicolumn{2}{c}{93.00 $\pm$ 1.05} \\
				FT-Transformer \cite{FTT}      & \multicolumn{2}{c|}{74.66 $\pm$ 1.25} & \multicolumn{2}{c}{94.30 $\pm$ 0.29} \\
				Concat Fuse \cite{CF}          & \multicolumn{2}{c|}{76.49 $\pm$ 0.61} & \multicolumn{2}{c}{90.37 $\pm$ 7.54} \\
				\midrule
				\multicolumn{5}{c}{\textbf{SSL Multimodal Pre-training Methods}} \\
				\midrule
				MMCL \cite{PTM2}               & 76.31 $\pm$ 0.45 & 77.18 $\pm$ 0.82 & 93.51 $\pm$ 1.06 & 97.45 $\pm$ 0.19 \\
				TIP \cite{PTM6}                & 73.02 $\pm$ 0.40 & 75.87 $\pm$ 0.44 & 98.84 $\pm$ 0.05 & 97.91 $\pm$ 0.11 \\
				AMF-MedIT (ours)               & 76.97 $\pm$ 0.75 & \textbf{78.36 $\pm$ 0.12} & 94.76 $\pm$ 0.45 & \textbf{98.89 $\pm$ 0.11} \\
				\bottomrule
			\end{tabular}
		\begin{tablenotes}
			\footnotesize
			\item[1] Frozen strategy.
			\item[2] Trainable strategy.
		\end{tablenotes}
	\end{threeparttable}
\end{table*}

The results are summarized in Table \ref{Multi-modal}. For the OL3I task, AMF-MedIT achieves the best performance under the trainable strategy, with an AUC of 78.36\%, outperforming both MMCL (77.18\%) and TIP (75.87\%). For the DVM task, which contains a substantially larger number of training samples, AMF-MedIT also achieves the highest accuracy of 98.89\%, marginally higher than TIP (98.84\%).
While TIP can fully exploit its attention-intensive architecture and achieve competitive performance when ample data are available, its reliance on large datasets limits its generalization to smaller medical cohorts. In contrast, AMF-MedIT shows consistent advantages across both large- and small-scale settings, benefiting from its straightforward yet adaptive fusion design. Furthermore, compared with supervised unimodal baselines, the improvements are substantial, underscoring the advantages of multimodal SSL pretraining.
\subsection{Robustness to Noisy Data}
\begin{figure*}[h]
	\centering  
	\subfigcapskip=-2pt
	\subfigbottomskip=-2pt
	\subfigure[Performance on missing tabular data and clean image data]{ 
		\includegraphics[width=0.45\textwidth]{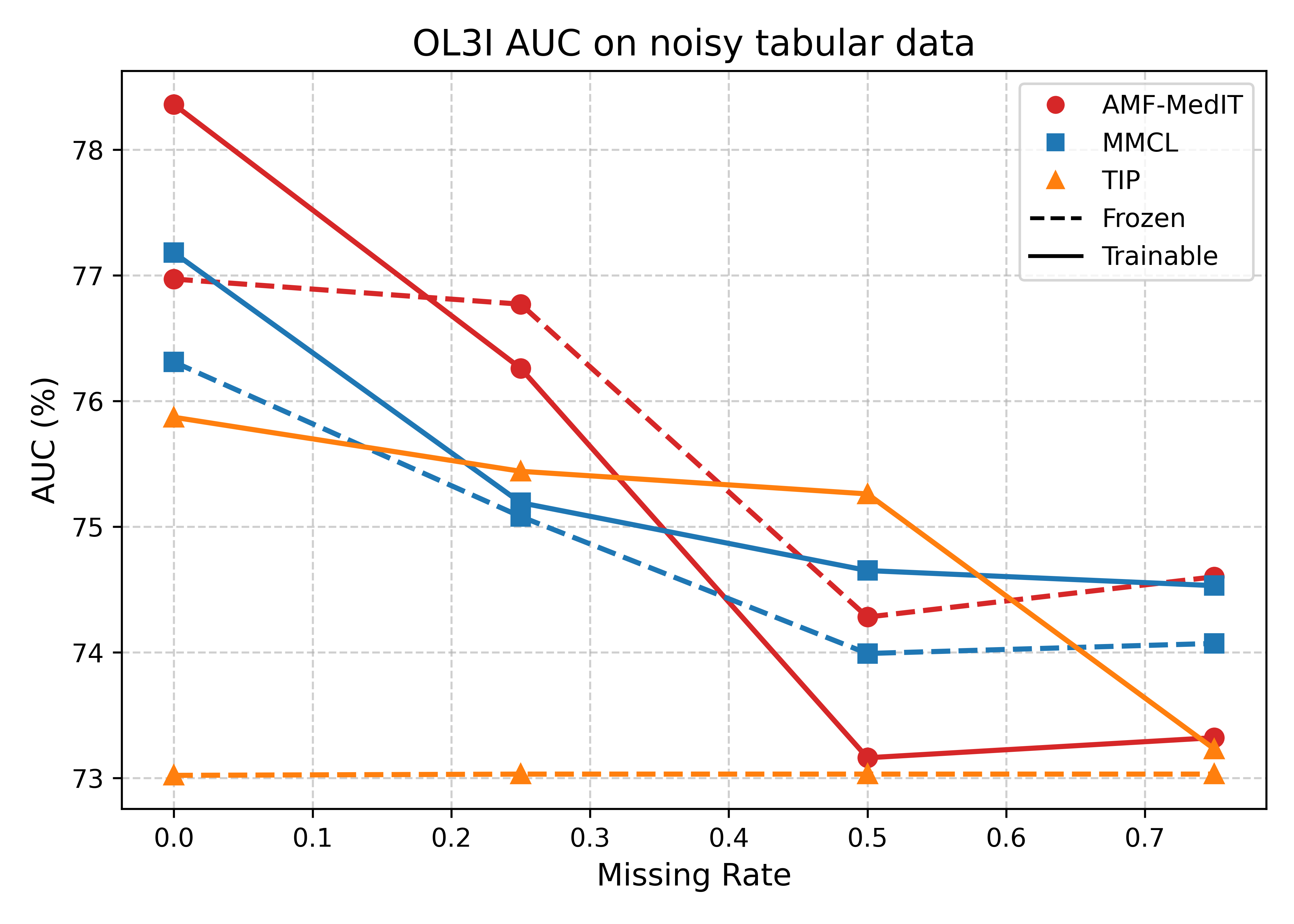}
		\label{noisy tabular}
	}
	\hfill
	\subfigure[Performance on noisy image data and clean tabular data]{ 
		\includegraphics[width=0.45\textwidth]{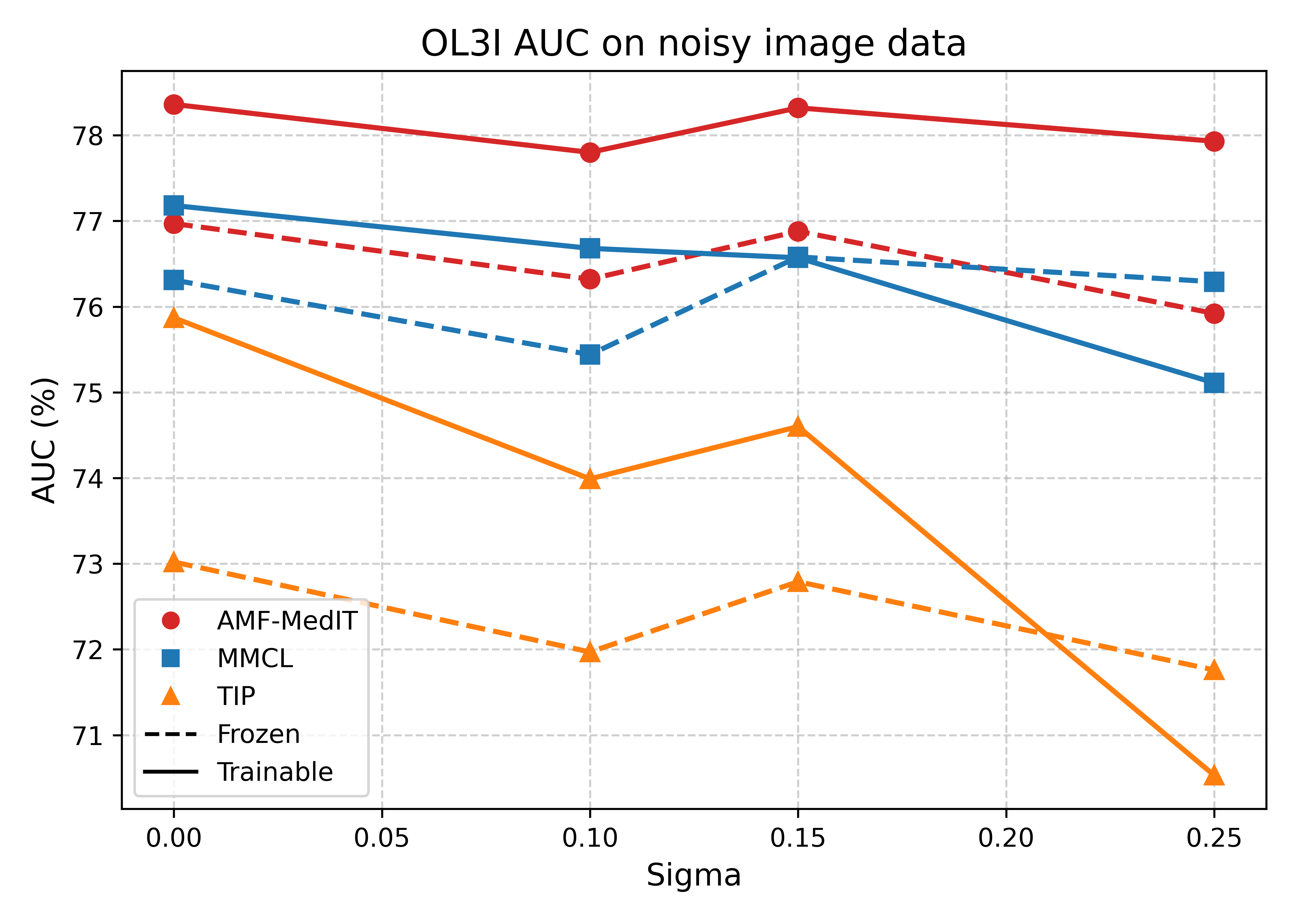}
		\label{noisy image}
	}
	\caption{Performance comparison with SSL image-tabular methods on noisy data.}
	\label{noisy data}
\end{figure*}

To evaluate the robustness of the models under noisy conditions, we simulated two types of noise commonly encountered in clinical practice based on the OL3I dataset: tabular data missingness and additive Gaussian noise in images.
\begin{itemize}
	\item Tabular data missingness: We adopted the random feature missingness (RFM) strategy, where a random subset of feature columns is removed and the missing values are imputed with the mean of the corresponding feature. The missing rates were set to $[0, 0.25, 0.5, 0.75]$.
	\item Image Gaussian noise: We injected additive Gaussian noise into the images, with noise levels controlled by the standard deviation $\sigma$, set to $[0, 0.1, 0.15, 0.25]$.
\end{itemize}
We designed two experimental scenarios: (i) noisy tabular data with clean images, and (ii) noisy images with clean tabular data. In each scenario, the noise was applied to the training, validation, and test splits of the downstream tasks, and we fine-tuned the pretrained models of three image-tabular SSL frameworks (MMCL \cite{PTM2}, TIP \cite{PTM6}, and ours) on the noisy datasets and compared their downstream performance.
Fig. \ref{noisy data} presents the multimodal performance on noisy data.

For tabular missingness, TIP achieves relatively low absolute AUC but shows the smallest performance drop, indicating strong stability. This advantage comes from its dedicated missing-feature processing mechanism, which masks absent tabular features and enables consistent adaptation under both frozen and trainable settings. 
Although AMF-MedIT demonstrates strong performance and good adaptability under the frozen setting, its performance degrades notably under the trainable setting when the missing rate exceeds 0.5. This limitation may arise because $r_{\text{conf}}$ is configured based on the frozen-encoder performance in our experiments; at high missing rates, the confidence scores of frozen and trainable encoders diverge, leading to suboptimal adjustment. 
In future work, the settings of $r_{\text{conf}}$ could be refined by leveraging unimodal performance under the trainable strategy to achieve a more accurate estimation, and even further adjusted by incorporating domain expertise from medical professionals.

For image noise, AMF-MedIT delivers the best overall performance with minimal degradation, demonstrating strong adaptability. TIP, however, is less effective in this case. Since its mechanism is tailored only for handling tabular missingness, it cannot accommodate degraded image quality, resulting in weaker robustness compared with AMF-MedIT and MMCL.

Overall, AMF-MedIT exhibits the strongest robustness across both noise scenarios. Rather than depending on a modality-specific mechanism, the $r_{\text{conf}}$ adjustment in our model operates in a modality-agnostic manner, providing broader adaptability.

\subsection{Ablation Studies}
\subsubsection{Ablation Studies on Tabular Encoder and Model Interpretation}
To evaluate the effectiveness of the proposed FT-Mamba tabular encoder within AMF-MedIT, we replace it with three representative tabular encoders: MLP \cite{PTM2, CF}, ResNet \cite{FTT}, and FT-Transformer \cite{FTT}. Each encoder is trained under three settings: (i) unimodal supervised training, (ii) multimodal contrastive pretraining, and (iii) multimodal fine-tuning on downstream tasks. To further interpret how different tabular encoders operate and influence contrastive learning, we additionally perform Integrated Gradients (IG) analysis \cite{IG} on unimodal tabular models and Grad-CAM analysis \cite{GradCAM} on multimodal encoders.

\paragraph{\textbf{Performance.}}
The results of tabular encoders on the OL3I and DVM datasets are summarized in Table \ref{tabular-encoders}. On the OL3I dataset, FT-Mamba achieves the best AUC both in unimodal (77.36\%) and multimodal (78.36\%) settings. On the DVM dataset, FT-Transformer achieves the best overall accuracy in both unimodal supervised training (94.30\%) and multimodal fine-tuning (99.23\%), with FT-Mamba ranking closely behind (92.71\% and 98.89\%, respectively). In contrast, MLP and ResNet performed much worse.
Overall, these results suggest that FT-Mamba offers better adaptability across datasets of different scales: it shows clear advantages on the smaller OL3I dataset, while remaining competitive with FT-Transformer on the larger DVM dataset. 
\begin{table*}[pos=h]
	\centering
	\caption{Ablation studies on different tabular encoders.}
	\label{tabular-encoders}
	\resizebox{\textwidth}{!}{
		\begin{tabular}{l | c | cc | c | cc}
			\toprule
			\multirow{2}{*}{\textbf{Tabular Encoder}} & 
			\multicolumn{3}{c|}{\textbf{OL3I AUC (\%)}} & 
			\multicolumn{3}{c}{\textbf{DVM ACC (\%)}} \\
			\cmidrule(lr){2-4} \cmidrule(lr){5-7}
			& Unimodal & Multimodal \faSnowflake & Multimodal \faFire* 
			& Unimodal & Multimodal \faSnowflake & Multimodal \faFire* \\
			\midrule
			MLP \cite{PTM2}             & 74.67 $\pm$ 2.06 & 76.63 $\pm$ 0.22 & 76.86 $\pm$ 0.35 
			& 85.50 $\pm$ 0.50 & 94.92 $\pm$ 0.43 & 97.88 $\pm$ 0.42 \\
			ResNet \cite{FTT}           & 74.43 $\pm$ 1.08 & 76.50 $\pm$ 0.83 & 75.46 $\pm$ 0.92 
			& 85.19 $\pm$ 0.37 & 95.09 $\pm$ 0.54 & 97.28 $\pm$ 0.70 \\
			FT-Transformer \cite{FTT}   & 74.66 $\pm$ 1.25 & 75.62 $\pm$ 0.49 & 76.46 $\pm$ 1.57 
			& \textbf{94.30 $\pm$ 0.29} & 93.83 $\pm$ 0.90 & \textbf{99.23 $\pm$ 0.35} \\
			FT-Mamba (ours)             & \textbf{77.36 $\pm$ 1.58} & 76.97 $\pm$ 0.75 & \textbf{78.36 $\pm$ 0.12} 
			& 92.71 $\pm$ 0.28 & 94.76 $\pm$ 0.45 & 98.89 $\pm$ 0.11 \\
			\bottomrule
		\end{tabular}
	}
	
\end{table*}

\begin{figure*}[pos=h]
	\centering  
	\subfigbottomskip=-2pt 
	\subfigcapskip=-2pt 
	\subfigure[MLP]{
		\includegraphics[width=0.23\linewidth]{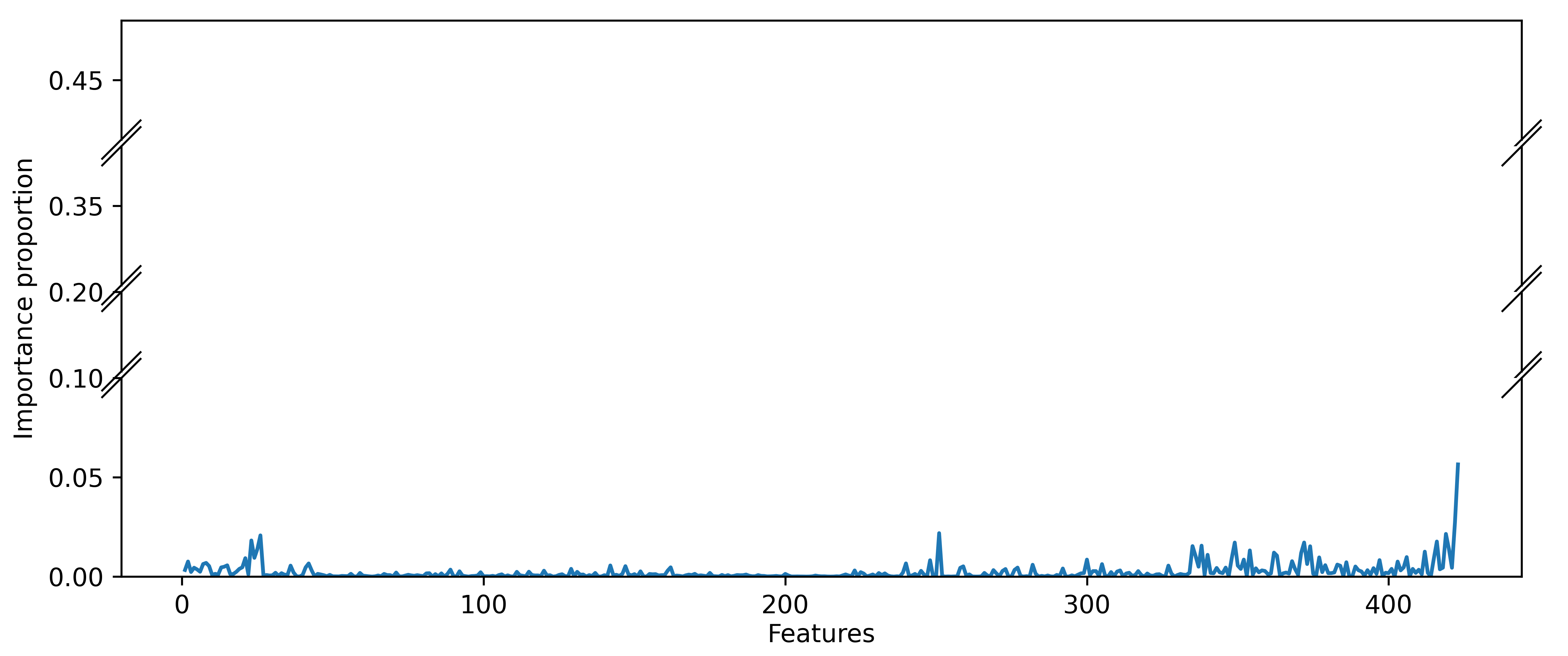} 
		\label{app-IG1a}}
	\subfigure[ResNet]{
		\includegraphics[width=0.23\linewidth]{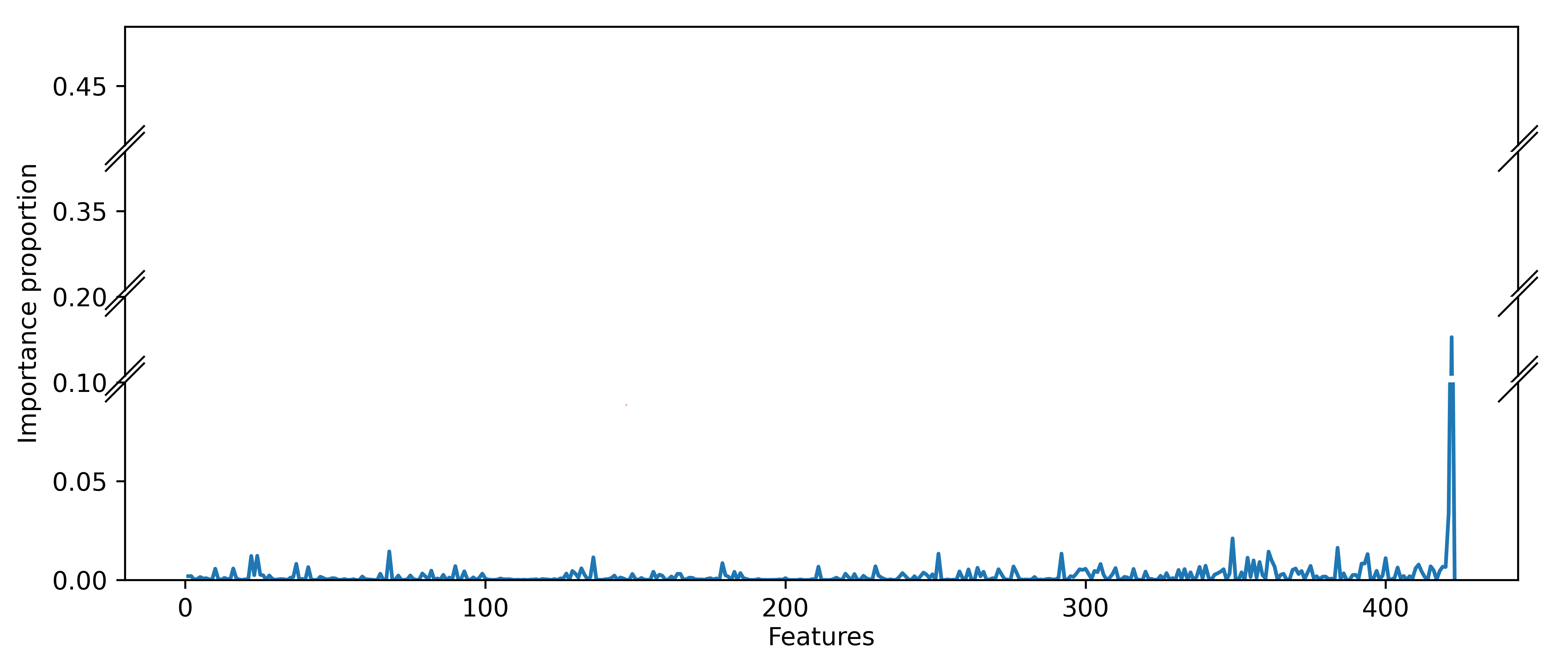}
		\label{app-IG1b}} 
	\subfigure[FT-Transformer]{
		\includegraphics[width=0.23\linewidth]{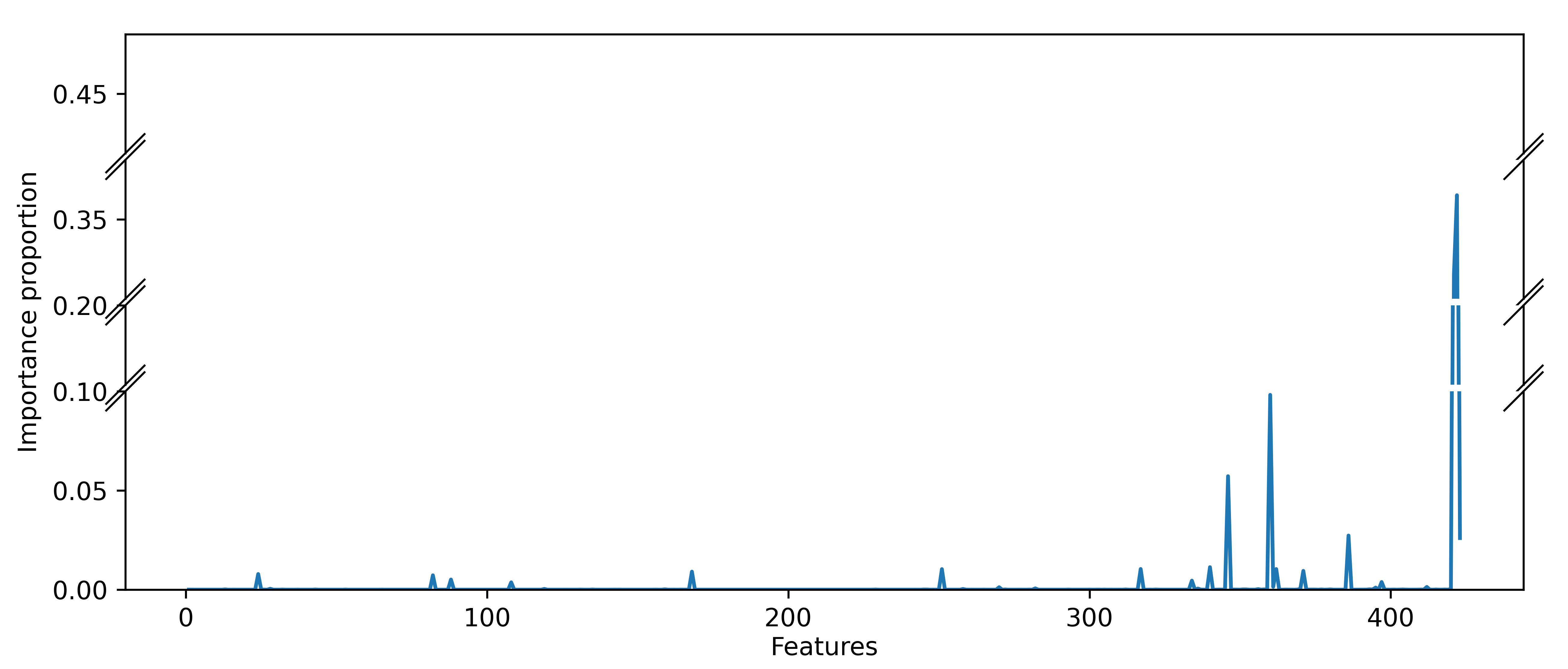}
		\label{app-IG1c}}
	\subfigure[FT-Mamba]{
		\includegraphics[width=0.23\linewidth]{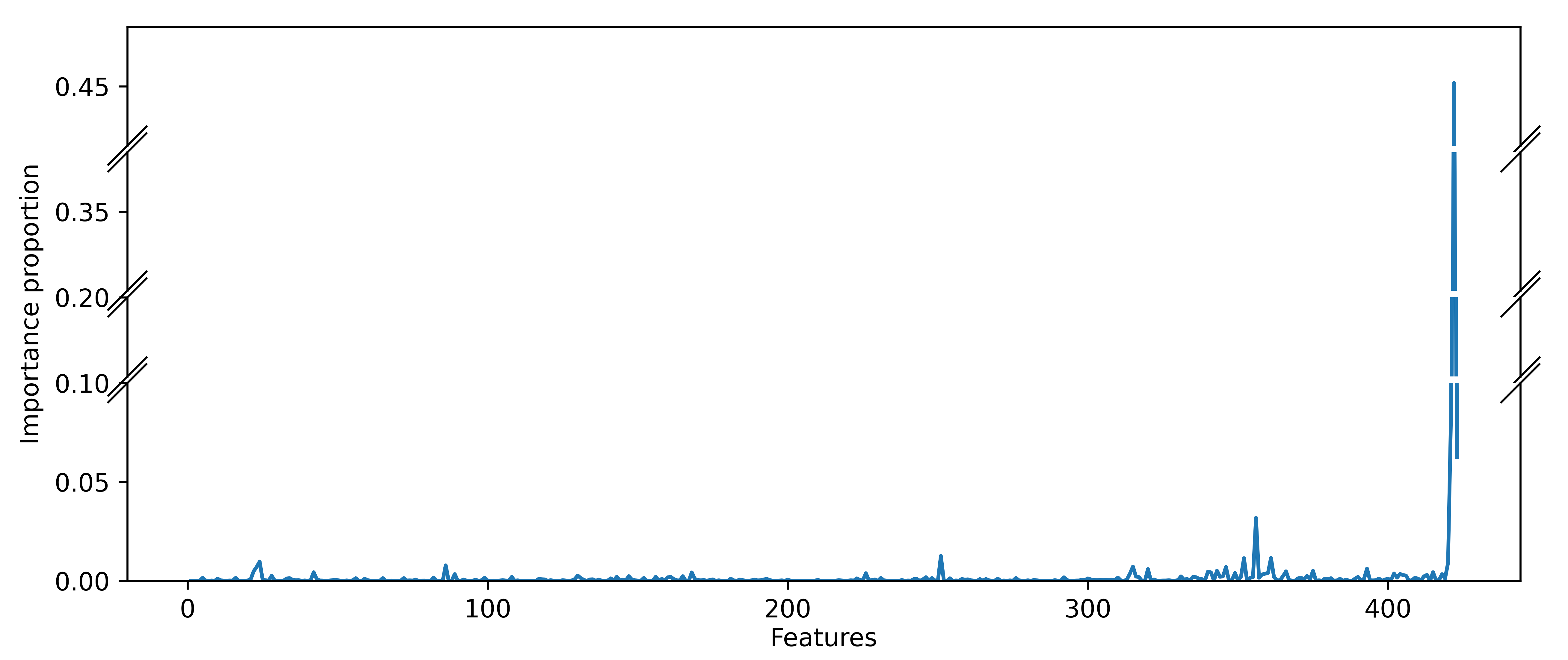}
		\label{app-IG1d}} 
	
	\caption{Importance distribution over 423 input features for unimodal tabular models. }
	\label{app-IG1}
\end{figure*}

\begin{figure*}[pos=h]
	\centering  
	\subfigcapskip=-2pt
	\subfigbottomskip=-2pt
	\subfigure[Importance proportions by category]{ 
		\includegraphics[width=0.27\textwidth]{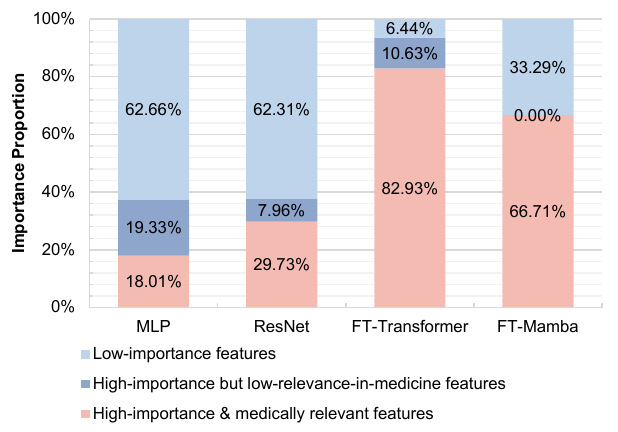}
		\label{tabularIG-2}
	}
	\hfill
	\subfigure[High-importance features of FT-Mamba]{ 
		\includegraphics[width=0.35\textwidth]{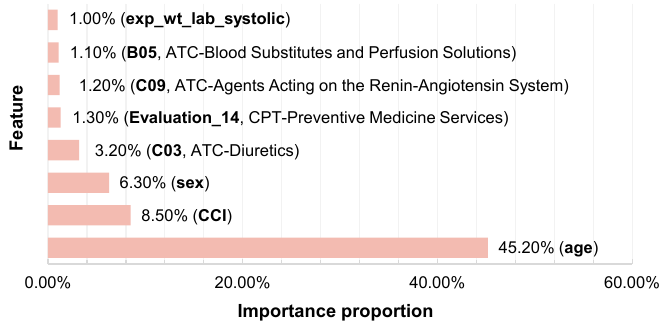}
		\label{tabularIG-3}
	}
	\hfill
	\subfigure[High-importance features of FT-Transformer]{ 
		\includegraphics[width=0.32\textwidth]{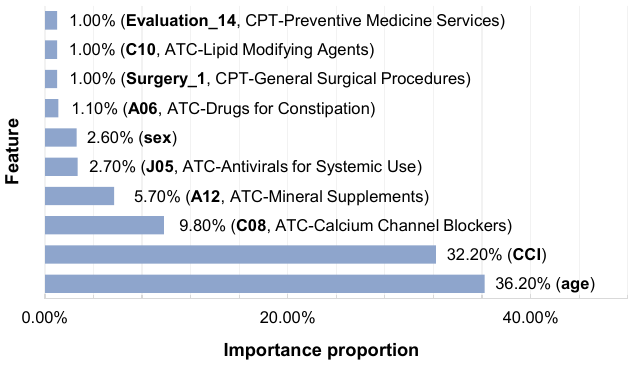}
		\label{tabularIG-4}
	}
	\caption{IG analysis results for unimodal tabular models. (a): Feature importance proportions by category for the OL3I dataset. Features with an importance percentage less than 1\% were classified as low-importance features (light blue). Features with an importance percentage  of 1\% or greater were classified as high-importance features. Based on \cite{PTMS16} and established medical knowledge, we further divided the high-importance features into medically relevant features (pink) and low-relevance features (dark blue).
		(b) and (c): High-importance features and their importance proportion for FT-Mamba and FT-Transformer. (exp\_wt\_lab\_systolic: exponentially weighted average of systolic blood pressure. CPT: Current Procedural Terminology codes, ATC: Anatomical Therapeutic Chemical codes.)}
\end{figure*}

\paragraph{\textbf{Integrated gradients analysis of tabular encoders.}}
Fig. \ref{app-IG1} shows the overall distribution of feature importance estimated by IG across all tabular features on the OL3I dataset, while Fig. \ref{tabularIG-2} groups the 423 features into three clinical categories for clarity. More detailed IG results are provided in Appendix \ref{AB}. The results reveal that MLP and ResNet show weak feature discrimination: MLP distributes attention almost uniformly across features, while ResNet exhibits a single peak but still spreads over many irrelevant variables. Quantitatively, their most important features account for only 18\% (ResNet) and even less for MLP of the total importance, far below FT-Mamba (45\%) and FT-Transformer (36\%) according to Table~\ref{app-IG2}. As a result, over 60\% of their attention remains dispersed across low-importance features, reflecting a limited ability to extract clinically meaningful signals. 

In contrast, FT-Mamba and FT-Transformer concentrate on a small subset of features, with more than 60\% of their attention massed on clinically salient variables. However, the patterns differ in quality. Fig. \ref{tabularIG-3} and \ref{tabularIG-4} show the specific high-importance features for FT-Mamba and FT-Transformer. FT-Mamba’s high-importance features are strongly aligned with well-established cardiovascular risk factors for IHD, such as age, sex, Charlson Comorbidity Index (CCI), and records related to the cardiovascular system, which are consistent with the prior findings \cite{PTMS16}. By comparison, FT-Transformer assigns substantial attention to features with little medical relevance (e.g., A12, A06, J05), suggesting it may capture noise patterns when training data are limited.

\begin{figure}[pos=h]
	\centering{\includegraphics[width=0.6\textwidth]{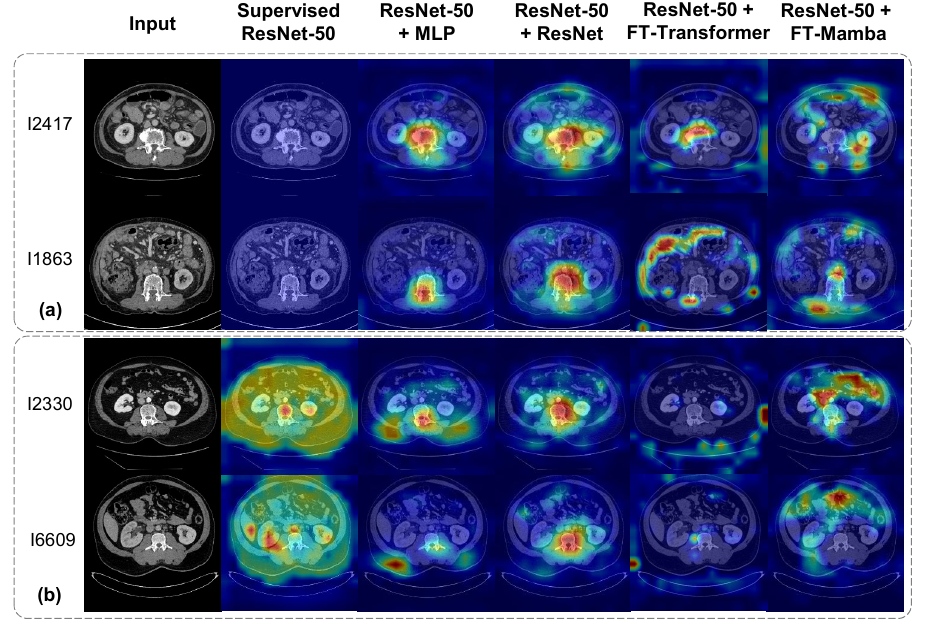}}
	\caption{CAMs generated by supervised ResNet-50 and pretrained multimodal models for IHD prediction. Group (a) illustrates activation patterns in non-IHD cases, while group (b) are IHD-positive samples.}
	\label{CAM}
\end{figure}

\paragraph{\textbf{Grad-CAM analysis of multimodal encoders.}}
Fig. \ref{CAM} visualizes Grad-CAMs from different image–tabular encoder pairs, alongside a unimodal supervised ResNet-50. The supervised ResNet-50 shows no distinct focal regions, confirming its limited capacity to capture meaningful visual cues under scarce supervision. Models with MLP or ResNet tabular encoders produce narrow attention localized around the lumbar spine, reflecting weak and overly singular guidance to the image encoder. In contrast, FT-Mamba and FT-Transformer enable broader, multi-point attention patterns, consistent with their stronger long-range modeling ability. However, FT-Transformer often activates irrelevant or blank areas, suggesting its noisier tabular features misguide the image encoder during contrastive training. By comparison, FT-Mamba directs attention toward clinically relevant structures—including musculature, subcutaneous and visceral fat, and cardiovascular regions—among which visceral fat proportion is a key biomarker in predicting IHD \cite{PTMS16}. 

Taken together, these results highlight the flexible and efficient capability of FT-Mamba in modeling tabular data. It can effectively extract highly relevant tabular features without being constrained by limited data or abundant irrelevant variables. Furthermore, it demonstrates superior capacity to provide medically meaningful supervision for the image modality. This simple yet effective working mechanism enables AMF-MedIT to achieve advantages in learning both tabular and image feature representations.

\subsubsection{Ablation Study on Multimodal Fusion Module}
To evaluate the effectiveness of the proposed AMF module, we replaced it with two mainstream multimodal fusion methods: concatenation and cross-attention. In the concatenation approach \cite{PTM2, CF}, the unimodal feature vectors generated by the pretrained encoders are concatenated and fed into a fully connected layer for classification. For cross-attention \cite{PTM6}, the feature map from the final ResNet-50 layer is flattened into image tokens, which are then combined with the tabular token generated by FT-Transformer or FT-Mamba and passed through a 2-layer multimodal cross-attention module with an embedding dimension of 512. The resulting CLS token is finally used for prediction. Table \ref{fusion-modulec} summarizes both classification performance and computational costs. All measurements were conducted with a batch size of 1. The inference latency was measured on an NVIDIA RTX 4090 GPU, averaged over 100 runs.
\begin{table*}[pos=h]
	\centering
	\caption{Ablation studies on different multimodal fusion modules.}
	\label{fusion-modulec}
	\resizebox{\textwidth}{!}{ 
		\begin{tabular}{l c | cc cc | cc cc}
			\toprule
			\multirow{2}{*}{\textbf{Fusion Module}} & 
			\multirow{2}{*}{\textbf{\#Param (M)}} &
			\multicolumn{2}{c}{\textbf{OL3I AUC (\%)}} & 
			\multicolumn{2}{c|}{\textbf{OL3I (Cost)}} &
			\multicolumn{2}{c}{\textbf{DVM ACC (\%)}} & 
			\multicolumn{2}{c}{\textbf{DVM (Cost)}} \\
			\cmidrule(lr){3-4} \cmidrule(lr){5-6} \cmidrule(lr){7-8} \cmidrule(lr){9-10}
			& & \faSnowflake & \faFire* & FLOPs & Lat. & \faSnowflake & \faFire* & FLOPs & Lat. \\
			\midrule
			Concact \cite{PTM2}         & -    & 75.80 $\pm$ 0.24 & 76.49 $\pm$ 0.60 & -      & -    & 89.08 $\pm$ 1.06 & 97.41 $\pm$ 0.36 & -      & -    \\
			Cross-attn. \cite{FTT}      & 9.49 & 73.36 $\pm$ 0.30 & 73.48 $\pm$ 0.98 & 3.67   & 1.54 & 96.96 $\pm$ 0.52 & 98.15 $\pm$ 0.31 & 3.17   & 1.21 \\
			AMF (ours)                  & 8.53 & 76.97 $\pm$ 0.75 & \textbf{78.36 $\pm$ 0.12} & 0.0085 & 0.12 & 94.76 $\pm$ 0.45 & \textbf{98.89 $\pm$ 0.11} & 0.0085 & 0.12 \\
			\bottomrule
		\end{tabular}
	}
	\begin{tablenotes}
		\footnotesize
		\item \textbf{Note:} FLOPs = floating-point operations (in GMac, $10^9$ multiply–accumulate operations). Lat. = latency (in ms). 
	\end{tablenotes}
\end{table*}

 On the OL3I dataset, AMF achieves the highest AUC (78.36\%) under the trainable strategy, outperforming concatenation (76.49\%) and cross-attention (73.48\%). On the DVM dataset, AMF remains competitive under a trainable strategy (98.89\%), demonstrating robust performance across datasets of different scales. These results confirm that simple fusion methods lack sufficient representational capacity, while highly complex methods like cross-attention are less robust under low-data regimes.

From an efficiency perspective, the cross-attention module incurs substantially higher parameters, FLOPs, and inference latency compared to AMF, reflecting its greater computational complexity. Notably, its cost grows quadratically with the spatial resolution of input images, due to token-level pairwise attention. In contrast, AMF performs interaction based on modality-level confidence scores rather than token-level pairwise computation, making its complexity effectively independent of image size. This design ensures that AMF is both scalable and practical, providing a strong balance between performance and efficiency for real-world multimodal applications.

\subsubsection{Ablation Study on Loss Functions}
To evaluate the contribution of the proposed auxiliary loss functions—magnitude loss and leakage loss, we fixed $r_{\text{conf}}=0.95$ and compared four different loss configurations during fine-tuning: 
(1) $\mathcal{L}_{\text{fine-tuning}} = \mathcal{L}_{\text{ce}}$; 
(2) $\mathcal{L}_{\text{fine-tuning}} = \mathcal{L}_{\text{ce}} + \lambda_1 \mathcal{L}_{\text{leakage}}$; 
(3) $\mathcal{L}_{\text{fine-tuning}} = \mathcal{L}_{\text{ce}} + \lambda_2 \mathcal{L}_{\text{magnitude}}$; 
(4) $\mathcal{L}_{\text{fine-tuning}} = \mathcal{L}_{\text{ce}} + \lambda_1 \mathcal{L}_{\text{leakage}} + \lambda_2 \mathcal{L}_{\text{magnitude}}$ (ours).
\begin{table}[pos=h]
	\centering
	\caption{AUCs of AMF-MedIT with different loss configurations on the OL3I dataset.}
	\label{loss_test}
		\begin{tabular}{ccc | cc | cc}
			\toprule
			\multirow{2}{*}{\textbf{$\mathcal{L}_{\text{ce}}$}} & 
			\multirow{2}{*}{\textbf{$\mathcal{L}_{\text{leakage}}$}} & 
			\multirow{2}{*}{\textbf{$\mathcal{L}_{\text{magnitude}}$}} & 
			\multicolumn{2}{c|}{\textbf{OL3I AUC (\%)}}  & \multicolumn{2}{c}{\textbf{DVM AUC (\%)}}\\
			\cmidrule(lr){4-5}\cmidrule(lr){6-7}
			& & & \faSnowflake & \faFire* & \faSnowflake & \faFire* \\
			\midrule
			\checkmark  & $\times$     & $\times$     & 73.11 $\pm$ 3.99 & 78.07 $\pm$ 0.64 & 88.31 $\pm$ 1.07 & 97.30 $\pm$ 0.36\\
			\checkmark  & \checkmark   & $\times$     & 75.46 $\pm$ 1.71 & 78.05 $\pm$ 0.36 & 91.01 $\pm$ 0.55 & 96.79 $\pm$ 0.70\\
			\checkmark  & $\times$     & \checkmark   & 74.11 $\pm$ 3.49 & 78.08 $\pm$ 0.35 & 91.20 $\pm$ 0.57 & 97.08 $\pm$ 0.30\\
			\checkmark  & \checkmark   & \checkmark   & \textbf{76.97 $\pm$ 0.75} & \textbf{78.36 $\pm$ 0.12} & \textbf{94.76 $\pm$ 0.45} & \textbf{98.89 $\pm$ 0.11}\\
			\bottomrule
		\end{tabular}
\end{table}

The AUCs under different loss configurations are presented in Table \ref{loss_test}. The best performance is achieved when all three losses are jointly applied.
Removing the leakage loss $\mathcal{L}_{\text{leakage}}$ leads to a substantial drop in performance, especially under the frozen strategy (-2.86\% AUC), indicating that the absence of information flow regulation across modalities causes significant semantic leakage and degraded fusion quality. Similarly, omitting the magnitude loss also reduces performance and introduces instability, as it fails to constrain the feature magnitude alignment between the image and tabular modalities, which can impede effective fusion.

The improvement brought by these two auxiliary losses is more evident in the frozen training setting, where the fusion module receives no gradient from the backbone. In such cases, relying solely on the classification loss $\mathcal{L}_{\text{ce}}$ is insufficient for guiding the training of the AMF module, leading to higher performance variance and lower overall accuracy.
These results collectively validate the necessity of both magnitude and leakage losses for effective multimodal training, as they enhance representation compatibility, stabilize optimization, and ultimately improve classification performance.

\section{Discussion}
In this work, we proposed AMF-MedIT, an efficient framework for integrating medical image and tabular data. Our approach was motivated by the substantial heterogeneity between modalities, particularly in terms of dimensionality and confidence levels, as well as the challenges posed by small and noisy clinical datasets. To address these issues, we designed a twofold strategy: (1) introducing prior knowledge through the AMF fusion module to achieve robust and adaptive integration across modalities, and (2) leveraging self-supervised contrastive pretraining together with an efficient tabular encoder, FT-Mamba, to enhance multimodal representation learning under limited data.

Compared with existing methods, AMF-MedIT demonstrates consistent advantages in both small-scale medical datasets and large-scale natural datasets. In clinical practice, multimodal learning often faces the dual challenges of data scarcity and noise. Medical image–tabular multimodal datasets remain limited in availability due to the high costs of data acquisition and the complexities of data management; moreover, missing clinical records or measurement errors can lead to unreliable tabular features, while artifacts and acquisition noise may compromise image quality. These challenges highlight that practical medical artificial intelligence (AI) systems require not only powerful models but also strategies that enhance efficiency and adaptability to limited data.
Our experiments show that AMF-MedIT strikes an effective balance between performance and data efficiency, benefiting from its simple yet adaptive design. Specifically, instead of relying on cross-attention-based modules, we adopt a modulation-based fusion strategy that incorporates prior knowledge, thereby alleviating the training burden and avoiding the data inefficiency often caused by overly complex fusion modules.   
Furthermore, thanks to its modality-agnostic confidence adjustment mechanism, AMF-MedIT exhibits robust tolerance to noise in both image and tabular modalities. In addition, the introduction of feature masks and leakage loss enables adaptive adjustment to changing downstream data conditions without requiring structural modifications to the network, thereby providing broader clinical applicability compared with methods tailored to specific noise types.

Ablation studies further highlight the key contributions of both the FT-Mamba encoder and the AMF module. 
First, the tabular encoder plays a central role in maintaining both high accuracy and data efficiency. Interpretability analyses reveal that FT-Mamba captures fine-grained, discriminative tabular features more effectively than MLP and ResNet encoders, which are constrained by local attention. Moreover, by employing a selective mechanism, it achieves efficient feature extraction that remains robust under limited data. Compared with transformer-based models that rely on full attention, it is capable of filtering out irrelevant features while preserving medically meaningful signals, without overfitting to spurious noise patterns introduced by model complexity. This capability also enhances contrastive pretraining and guides the image encoder toward learning precise and clinically relevant attention patterns. 
Second, the AMF module achieves strong performance with remarkably low computational complexity, in contrast to concatenation- or attention-based approaches. Importantly, its complexity is independent of image resolution, which is advantageous for small datasets and satisfies the real-time requirements of clinical applications. Finally, loss function ablations confirm that leakage loss and density loss contribute to training stability and overall performance.

Despite the observed benefits, our study has several limitations. First, while FT-Mamba provides a good trade-off under limited data, it still underperforms FT-Transformer on large-scale datasets, suggesting that FT-Transformer may be preferable as larger medical datasets become available. Second, the configuration of the $r_{\text{conf}}$ parameter in the AMF module remains sensitive. Our experiments on missing tabular features indicate that estimates derived from frozen strategies may not generalize well to trainable settings. More precise tuning strategies, potentially incorporating expert knowledge, are needed to improve robustness. Third, our validation was restricted to two multimodal datasets, and further evaluation on a wider range of clinical data is necessary. Lastly, our framework was applied only to classification tasks; its potential for other medical applications, such as segmentation or automated report generation, remains unexplored.

Looking forward, several directions can be pursued. Future work could focus on developing automated procedures to optimize $r_{\text{conf}}$ based on dataset-specific characteristics, while integrating domain expertise to guide its estimation. Extending the evaluation to more diverse modalities, such as MRI and ultrasound, will further establish the generalizability of our framework. Moreover, expanding AMF-MedIT beyond classification to support segmentation and report generation tasks could broaden its impact on multimodal medical AI.

\section{Conclusion}
In conclusion, we proposed AMF-MedIT for integrating medical image and tabular data.
At its core, the AMF module leverages streamlined modulation and prior knowledge to coordinate cross-modal disparities, providing an accurate and adaptable fusion strategy across diverse downstream conditions. Additionally, FT-Mamba serves as an efficient yet powerful tabular encoder, achieving a favorable balance between accuracy and data efficiency. Our interpretability analyses further reveal that encoder selection critically shapes multimodal contrastive pretraining, highlighting the essential role of FT-Mamba in multimodal feature representations.
Beyond architectural contributions, AMF-MedIT improves robustness against noise and reduces reliance on large annotated datasets, offering a practical and scalable solution to the realities of clinical multimodal learning. We hope this framework will inspire advancements in medical multimodal models and their applications in healthcare.

\section*{CRediT Authorship Contribution Statement}
\textbf{Congjing Yu:} Conceptualization, Data curation, Methodology, Validation, Writing - original draft, Writing - review \& editing. 
\textbf{Jing Ye:} Validation, Writing - review \& editing. 
\textbf{Yang Liu:} Validation, Writing - review \& editing. 
\textbf{Xiaodong Zhang:} Supervision.
\textbf{Zhiyong Zhang:} Funding acquisition, Supervision, Writing - review \& editing.

\section*{Declaration of Competing Interest}
The authors declare that they have no known competing financial interests or personal relationships that could have appeared to influence the work reported in this paper.

\section*{Acknowledgments}
The work was supported by the Science and Technology Planning Project of Key Laboratory of Advanced IntelliSense Technology, Guangdong Science and Technology Department (2023B1212060024).

\begin{appendices}
	\section{{Implementation Details}}
	\label{AA}
	\setcounter{figure}{0}
	\setcounter{table}{0}
	\renewcommand{\thefigure}{A.\arabic{figure}}
	\renewcommand{\theequation}{A.\arabic{equation}}
	\renewcommand{\thetable}{A.\arabic{table}}
	\subsection{Data Preprocessing}
	\label{AA1}
	For CT images in the OL3I dataset, the extreme grayscale values (greater than 275 and less than -125) were removed to enhance the image contrast. Then, the pixel values of images from both datasets were normalized to the range of 0 to 1. 
	
	For missing tabular data in the OL3I dataset, we applied median imputation. For tabular data in the DVM dataset, 13 features including 9 continuous features (Adv\_year, Adv\_month, Reg\_year, Runned\_Miles, Price, Seat\_num, Door\_num, Entry\_price, Engine\_size) and 4 categorical features (Color, Bodytype, Gearbox, Fuel\_type) were remained for input. The continuous tabular features in both datasets were normalized using z-score normalization before being fed into the model. 
	
	\subsection{Multimodal Pretraining}
	\label{AA2}
	The pretraining procedures and augmentations followed those described in \cite{PTM2}, where image augmentations were applied with a probability of 0.95 and tabular features were corrupted at a rate of 0.3. The optimizer used was Adam with a weight decay of $1e^{-4}$. The learning rate was selected based on validation performance, with possible values of $\{3e^{-3}, 3e^{-4}, 3e^{-4}\}$. The model was trained for 500 epochs using a cosine annealing scheduler with a warmup period of 10 epochs.
	
	\subsection{Downstream training}
	\label{AA3}
	Considering the severe class imbalance in the OL3I dataset, where class 0 contains 7,784 samples and class 1 only 355 samples, the weight factor $\alpha$ in the weighted CE loss was set to 0.1 for class 0 and 2 for class 1. Additionally, a weighted sampler strategy was applied to the OL3I dataset, where class 0 in the training set was downsampled by a factor of 0.5, and class 1 was oversampled by a factor of 5. 
	
	During the fine-tuning of the AMF module, $\lambda_1$ and $\lambda_2$ were empirically set to 5.0 to ensure that the magnitude and leakage losses have a comparable magnitude and regularization strength to the cross-entropy loss.
	
	To determine the embedding dimension $D_{tab}$ for tabular encoders, we evaluated multiple candidate dimensions during supervised training on the OL3I dataset. For MLP and ResNet, the values tested were $\{128, 256, 512, 1024\}$, and for FT-Transformer and FT-Mamba, $\{32, 64, 96, 128, 256, 384\}$ were considered. Based on hyperparameter search, the tabular embedding dimension $D_{tab}$ was set to 256 for MLP and ResNet, and 64 for FT-Transformer and FT-Mamba. The number of encoder blocks was fixed to 2 across all tabular models, following the configurations in \cite{PTM2, FTT}. 
	
	To optimize the training hyperparameters, we tested three learning rates $\{1e^{-2}, 1e^{-3}, 1e^{-4}\}$ using a fixed seed of 2022 for all supervised and fine-tuning experiments. In the fine-tuning of the cross-attention module, we observed severe overfitting and thus tested weight decay values of $\{1e^{-1}, 1e^{-2}, 1e^{-3}, 1e^{-4}\}$ to regularize training. In all other cases, the weight decay factor was fixed to 0. The optimal values were selected based on the validation performance.
	
	All experiments utilized the Adam optimizer. Early stopping was applied based on the validation metric, triggered if no improvement greater than 0.0002 was observed over 10 consecutive epochs. After determining the optimal hyperparameters, each model was retrained using five random seeds (2022–2026) to mitigate randomness.
	
	\subsection{Quantifying the Modality Confidence Ratio}
	\label{AA4}
	In our experiments, we assessed the confidence levels of unimodal feature vectors based on their respective unimodal performance. Specifically, after multimodal pretraining, we constructed unimodal models by utilizing the pretrained encoders and fine-tuned them on downstream tasks with a frozen strategy. The modality confidence ratio $r_{\text{conf}}$ was then calculated as the ratio of the image-to-tabular unimodal model metrics on the testing set. The unimodal performance and the corresponding modality confidence ratio $r_{\text{conf}}$ are presented in Table \ref{app-conf} and \ref{app-conf2}.
	\begin{table}[pos=h]
		\centering
		\caption{Unimodal encoder performance and corresponding modality confidence ratios with different tabular encoders.}
		\label{app-conf}
		\begin{tabular}{l|ccc | ccc}
			\toprule
			\multirow{2}{*}{\textbf{Tabular Encoder}} 
			& \multicolumn{3}{c|}{\textbf{OL3I}} 
			& \multicolumn{3}{c}{\textbf{DVM}} \\
			\cmidrule(lr){2-4} \cmidrule(lr){5-7}
			& Image AUC (\%) & Tabular AUC (\%) & $r_{\text{conf}}$ 
			& Image ACC (\%) & Tabular ACC (\%) & $r_{\text{conf}}$ \\
			\midrule
			MLP            & 74.99 & 77.41 & 0.97 & 66.11 & 79.65 & 0.83 \\
			ResNet         & 74.99 & 75.64 & 1.00 & 45.23 & 77.95 & 0.58 \\
			FT-Transformer & 75.62 & 74.62 & 1.01 & 34.20 & 71.23 & 0.48 \\
			FT-Mamba       & 72.92 & 76.21 & 0.95 & 60.20 & 66.78 & 0.90 \\
			\bottomrule
		\end{tabular}
	\end{table}
	\begin{table}[pos=h]
		\centering
		\caption{Unimodal performance of pretrained encoders and modality confidence ratios under noisy conditions.}
		\label{app-conf2}
		\begin{tabular}{cccc | cccc}
			\toprule
			\multirow{2}{*}{\textbf{Tabular Missing Rate}} 
			& \multicolumn{3}{c|}{\textbf{OL3I AUC (\%)}} 
			& \multirow{2}{*}{\textbf{Image $\sigma$}} 
			& \multicolumn{3}{c}{\textbf{OL3I AUC (\%)}} \\
			\cmidrule(lr){2-4} \cmidrule(lr){6-8}
			& Image & Tabular & $r_{\text{conf}}$ 
			& & Image & Tabular & $r_{\text{conf}}$ \\
			\midrule
			0.25 & 72.92 & 76.14 & 0.96 & 0.10 & 69.37 & 76.21 & 0.91 \\
			0.50 & 72.92 & 66.93 & 1.09 & 0.15 & 64.16 & 76.21 & 0.96 \\
			0.75 & 72.92 & 64.07 & 1.14 & 0.25 & 56.89 & 76.21 & 0.89 \\
			\bottomrule
		\end{tabular}
	\end{table}
	
	\subsection{Model interpretation}
	\label{AA5}
	IG estimates the contribution of each input feature by integrating gradient values along a straight-line path from a baseline input to the target sample \cite{IG}.  
	In our setting, the baseline input was defined as the mean of all training tabular samples. IG scores were computed for all training samples with respect to the disease class, and the final feature importance values were averaged to mitigate randomness. For consistent comparison, each feature’s importance was normalized as a percentage of the total importance within the model, highlighting its relative influence during inference.
	We also applied Grad-CAM \cite{GradCAM} to generate class activation maps (CAMs) from the pretrained multimodal models. For a consistent comparison, all models used a concatenation-based fusion framework to avoid confounding effects from fusion modules.

	\section{Detailed Results of Integrated Gradient Analysis}
	\label{AB}
	\setcounter{figure}{0}
	\setcounter{table}{0}
	\renewcommand{\thefigure}{B.\arabic{figure}}
	\renewcommand{\theequation}{B.\arabic{equation}}
	\renewcommand{\thetable}{B.\arabic{table}}
	
	Table \ref{app-IG2} presents the top 10 most important tabular features and their corresponding importance scores as identified by the IG analysis for the four unimodal tabular models. While MLP, ResNet, and FT-Transformer tend to focus on several variables that are less relevant to IHD, FT-Mamba demonstrates a strong emphasis on features closely associated with the cardiovascular system.
	\begin{table}[pos=h]
		\centering
		\caption{Top 10 most important tabular features for MLP, ResNet, FT-Transformer and FT-Mamba unimodal tabular models.}
		\label{app-IG2}
		\resizebox{\linewidth}{!}{
			\begin{tabular}{l|l|l|l}
				
				\hline
				\textbf{Tabular Model} & \textbf{Feature}       & \textbf{Description} & \textbf{Importance score}    \\ \hline
				\multirow{10}{*}{MLP} &
				gender                  & Whether the patient is male          & 0.056  \\ 
				&age\_at\_scan          & Age at time of abdominal scan in years          & 0.027  \\ 
				&Evaluation\_14         & Number of times a CPT code was documented in the group: Preventive Medicine Services          & 0.022  \\ 
				&V04                    & Number of prescriptions that match ATC code classification: Diagnostic Agents          & 0.021  \\  
				&smoker                 & Whether it was documented that the patient is a smoker in the year prior to abdominal scan          & 0.021  \\ 
				&exp\_wt\_lab\_diastolic& Exponentially weighted average of measurements diastolic blood pressure during the year prior to abdominal scan [mm Hg]          & 0.018  \\  
				&S02                    & Number of prescriptions that match ATC code classification: Otologicals          & 0.018  \\ 
				&D10                    & Number of prescriptions that match ATC code classification: Anti-Acne Preparations          & 0.017  \\ 
				&B01                    & Number of prescriptions that match ATC code classification: Antithrombotic Agents          & 0.017 \\  
				&A04                    & Number of prescriptions that match ATC code classification: Antiemetics And Antinauseants & 0.016 \\  \hline
				\multirow{10}{*}{ResNet} & 
				age\_at\_scan          & Age at time of abdominal scan in years          & 0.180  \\
				&CCI                    & Charlson Comorbidity Index          & 0.034    \\
				&B01                    & Number of prescriptions that match ATC code classification: Antithrombotic Agents & 0.021\\
				&J02                    & Number of prescriptions that match ATC code classification: Antimycotics For Systemic Use          & 0.016               \\
				&Chapter\_II\_4         & Number of times an ICD was documented in the year prior to imaging: Neoplasms of uncertain or unknown behaviour          & 0.014               \\ 
				&C09                    & Number of prescriptions that match ATC code classification: Agents Acting On The Renin-Angiotensin System          & 0.014               \\
				&Medical\_9             & Number of times a CPT code was documented in the group: Special Otorhinolaryngologic Services and Procedures           & 0.013               \\ 
				&Evaluation\_14         & Number of times a Current Procedural Terminology code was documented in the group: Preventive Medicine Services          & 0.013               \\ 
				&M02                    & Number of prescriptions that match ATC code classification: Topical Products For Joint And Muscular Pain         & 0.013               \\ 
				&exp\_wt\_lab\_systolic & Exponentially weighted average of measurements of systolic blood pressure during the year prior to abdominal scan [mm Hg]          & 0.012      \\ \hline
				\multirow{10}{*}{FT-Transformer}  &         
				age\_at\_scan    & Age at time of abdominal scan in years          & 0.362  \\ 
				&CCI              & Charlson Comorbidity Index          & 0.322  \\ 
				&C08              & Number of prescriptions that match ATC code classification: Calcium Channel Blockers         & 0.098  \\ 
				&A12              & Number of prescriptions that match ATC code classification: Mineral Supplements         & 0.057  \\  
				&J05              & Number of prescriptions that match ATC code classification: Antivirals For Systemic Use         & 0.027  \\ 
				&gender           & Whether the patient is male          & 0.026  \\  
				&A06              & Number of prescriptions that match ATC code classification: Drugs For Constipation         & 0.011  \\ 
				&Surgery\_1       & Number of times a CPT code was documented in the group: General Surgical Procedures          & 0.010  \\ 
				&C10              & Number of prescriptions that match ATC code classification: Lipid Modifying Agents         & 0.010 \\  
				&Evaluation\_14   & Number of times a CPT code was documented in the group: Preventive Medicine Services          & 0.010 \\ \hline              
				\multirow{10}{*}{FT-Mamba} &
				age\_at\_scan    & Age at time of abdominal scan in years          & 0.452  \\ 
				&CCI              & Charlson Comorbidity Index          & 0.085  \\ 
				&gender           & Whether the patient is male          & 0.063  \\ 
				&C03              & Number of prescriptions that match ATC code classification: Diuretics         & 0.032  \\  
				&Evaluation\_14   & Number of times a CPT code was documented in the group: Preventive Medicine Services          & 0.013  \\ 
				&C09              & Number of prescriptions that match ATC code classification: Agents Acting On The Renin-Angiotensin System         & 0.012  \\  
				&B05              & Number of prescriptions that match ATC code classification: Blood Substitutes And Perfusion Solutions         & 0.011  \\ 
				&exp\_wt\_lab\_systolic & Exponentially weighted average of measurements of systolic blood pressure during the year prior to abdominal scan [mm Hg]          & 0.010  \\ 
				&V06              & Number of prescriptions that match ATC code classification: General Nutrients         & 0.009 \\  
				&Chapter\_IX\_3   & Number of times an ICD was documented in the year prior to imaging: Hypertensive diseases          & 0.008 \\  \hline      
		\end{tabular}}
	\end{table}
\end{appendices}

\bibliography{reference}

\end{document}